\documentclass[10pt,twocolumn,letterpaper]{article}

\usepackage[pagenumbers]{cvpr} %

\usepackage{pgfplotstable}
\usepackage{adjustbox}
\usepackage{booktabs}
\usepackage{tikz}
\usetikzlibrary{calc}
\usepackage{bm}
\usetikzlibrary{backgrounds}
\usetikzlibrary{positioning} 
\usetikzlibrary{fit}
\usepackage{pgfplots}
\pgfplotsset{compat=1.18} 
\usepackage{graphicx}
\usepackage{verbatim}

\usepackage[accsupp]{axessibility}  %

\usepackage{tabularx}
\usepackage{multirow}
\usepackage{subcaption}
\usepackage{xcolor}

\newcommand\blfootnote[1]{
  \begingroup
  \renewcommand\thefootnote{}\footnote{#1}
  \addtocounter{footnote}{-1}
  \endgroup
}

\definecolor{cvprblue}{rgb}{0.21,0.49,0.74}
\colorlet{cvprblue_light}{cvprblue!50!white} %

\definecolor{lightgrey}{gray}{0.6}

\newcolumntype{Y}{>{\centering\arraybackslash}X}

\newcommand{\downrightarrow}{\hspace{0pt}%
  \raisebox{1.5pt}{\begin{tikzpicture}[scale=0.4, baseline=(current bounding box.south)]
    \draw[-latex] (0.0, 0.5) -- (0.0, 0.25) -- (0.55, 0.25);
  \end{tikzpicture}} %
}

\newcommand{\fancyrightarrow}{\hspace{0pt}%
  \raisebox{2.5pt}{\begin{tikzpicture}[scale=0.4, baseline=(current bounding box.south)]
    \draw[-latex] (0.0, 0.5) --  (0.55, 0.5);
  \end{tikzpicture}} %
}

\definecolor{c_red}{HTML}{ea4335}
\definecolor{c_blue}{HTML}{4285f4}
\definecolor{c_orange}{HTML}{ff6d01}
\definecolor{c_green}{HTML}{34a853}
\definecolor{c_purple}{HTML}{9900ff}
\definecolor{c_yellow}{RGB}{241,191,66}
\definecolor{c_other}{HTML}{000000}

\colorlet{tablered}{c_red!99!black}
\colorlet{tablegreen}{c_green!99!black}

\newcommand{\qincnew}[1]{\textcolor{tablegreen}{\tiny \textsuperscript{$\uparrow$#1}}}
\newcommand{\qdecnew}[1]{\textcolor{tablered}{\tiny \textsuperscript{$\downarrow$#1}}}

\makeatletter
\newcommand\notsotiny{\@setfontsize\notsotiny{6.5}{7.5}}
\newcommand\notscriptsize{\@setfontsize\notscriptsize{7.5}{8.5}}
\makeatother

\usepackage{pifont}
\newcommand{\cmark}{\ding{51}}
\newcommand{\xmark}{\ding{55}}
\newcommand{\rcmark}{\textcolor{tablered}{\cmark}}

\newcommand{\gxmark}{\textcolor{tablegreen}{\xmark}}

\newcommand{\tablestep}[1]{\textcolor{gray!90}{(#1)}}

    \colorlet{adapterlinecolor}{cvprblue}
    \colorlet{eomtcolor}{c_orange}
    \colorlet{eomtmaskedcolor}{eomtcolor!50!white}

    \colorlet{vitoutlinecolor}{eomtcolor} %
    \colorlet{vitfillcolor}{vitoutlinecolor!10!white} %
    \colorlet{vitopfillcolor}{vitfillcolor!87!vitoutlinecolor} %
    \colorlet{newblocksoutlinecolor}{c_green!90!white} %
    \colorlet{newblocksfillcolor}{newblocksoutlinecolor!10!white} %
    \colorlet{newblocksopfillcolor}{newblocksfillcolor!70!newblocksoutlinecolor} %
    
    \colorlet{outputoutlinecolor}{c_blue!70!black!50!white}
    \colorlet{outputfillcolor}{c_blue!70!black!10!white}

    \newcommand{\annealingplotsamples}{60}
    \def\maskoutcol{black!70!white}

    \definecolor{queryquerycol}{HTML}{fff9d6}
    \definecolor{querypatchcol}{HTML}{D6E8FF}
    \definecolor{patchquerycol}{HTML}{fcc5c5}
    \definecolor{patchpatchcol}{HTML}{deffd6}
    \definecolor{queryquerycoldark}{HTML}{f5d000}
    \definecolor{querypatchcoldark}{HTML}{4C8DFF}
    \definecolor{patchquerycoldark}{HTML}{e90c0c}
    \definecolor{patchpatchcoldark}{HTML}{29cc00}
    
    \colorlet{queryquerycol}{c_yellow!22!white}
    \colorlet{querypatchcol}{c_blue!22!white}
    \colorlet{patchquerycol}{c_red!22!white}
    \colorlet{patchpatchcol}{c_green!22!white}
    \colorlet{queryquerycoldark}{c_yellow}
    \colorlet{querypatchcoldark}{c_blue}
    \colorlet{patchquerycoldark}{c_red}
    \colorlet{patchpatchcoldark}{c_green}

    \definecolor{annealinglayertwentyone}{RGB}{62,7,18}
    \definecolor{annealinglayertwentytwo}{RGB}{36,102,138}
    \definecolor{annealinglayertwentythree}{RGB}{98,182,123}
    \definecolor{annealinglayertwentyfour}{RGB}{249,232,85}

\newcommand{\PAR}[1]{\vskip4pt \noindent {\bf #1~}}
\newcommand{\PARbegin}[1]{\noindent {\bf #1~}}

\expandafter\def\expandafter\normalsize\expandafter{%
    \normalsize%
    \setlength\abovedisplayskip{2pt}%
    \setlength\belowdisplayskip{2pt}%
    \setlength\abovedisplayshortskip{2pt}%
    \setlength\belowdisplayshortskip{2pt}%
}

\usepackage[pagebackref,breaklinks,colorlinks,allcolors=cvprblue]{hyperref}

\def\paperID{2093} %
\def\confName{CVPR}
\def\confYear{2025}

\title{Your ViT is Secretly an Image Segmentation Model}

\author{
Tommie Kerssies\textsuperscript{1} \qquad 
Niccolò Cavagnero\textsuperscript{2,*} \qquad 
Alexander Hermans\textsuperscript{3} \qquad 
Narges Norouzi\textsuperscript{1}\\
Giuseppe Averta\textsuperscript{2} \qquad 
Bastian Leibe\textsuperscript{3} \qquad 
Gijs Dubbelman\textsuperscript{1} \qquad 
Daan de Geus\textsuperscript{1,3}\\[0.6em]
$^1$Eindhoven University of Technology\qquad $^2$Polytechnic of Turin \qquad $^3$RWTH Aachen University\\
}

\begin{document}
\maketitle
\begin{abstract}
Vision Transformers (ViTs) have shown remarkable performance and scalability across various computer vision tasks. To apply single-scale ViTs to image segmentation, existing methods adopt a convolutional adapter to generate multi-scale features, a pixel decoder to fuse these features, and a Transformer decoder that uses the fused features to make predictions. In this paper, we show that the inductive biases introduced by these task-specific components can instead be learned by the ViT itself, given sufficiently large models and extensive pre-training. Based on these findings, we introduce the Encoder-only Mask Transformer (EoMT), which repurposes the plain ViT architecture to conduct image segmentation. With large-scale models and pre-training, EoMT obtains a segmentation accuracy similar to state-of-the-art models that use task-specific components. At the same time, EoMT is significantly faster than these methods due to its architectural simplicity, \eg, up to $4\times$ faster with \mbox{ViT-L}. Across a range of model sizes, EoMT demonstrates an optimal balance between segmentation accuracy and prediction speed, suggesting that compute resources are better spent on scaling the ViT itself rather than adding architectural complexity. Code: \href{https://www.tue-mps.org/eomt/}{https://www.tue-mps.org/eomt/}.
\end{abstract}
 
\blfootnote{\textsuperscript{*} Work done while visiting RWTH.}
\section{Introduction}
\label{sec:intro}

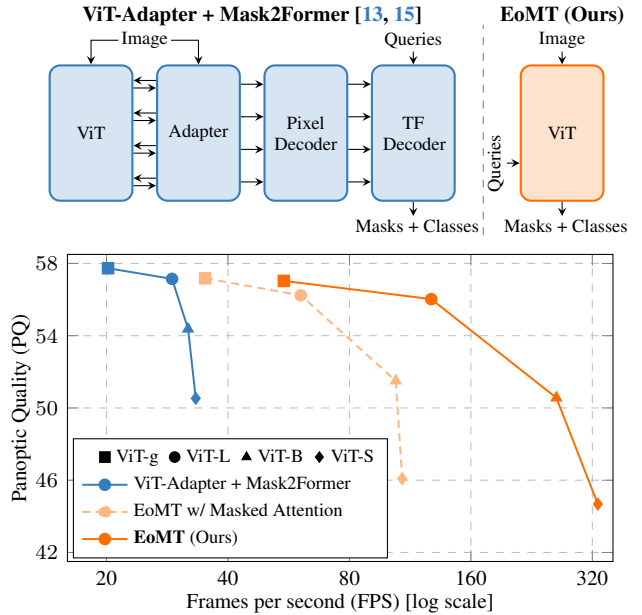
\begin{figure}[t]
    \begin{tikzpicture}[tight background]        
        \def\vitgmark{square*}
        \def\vitLmark{*}
        \def\vitBmark{triangle*}
        \def\vitSmark{diamond*}

        \def\vitadaptergdata{(20.1752915351241, 57.7339161459988)}
        \def\vitadapterLdata{(29.0763704139541, 57.1447821436729)}
        \def\vitadapterBdata{(31.8561473718594, 54.3710140968164)}
        \def\vitadapterSdata{(33.2832989527359, 50.5324811995185)}

        \def\eomtmaskedgdata{(35.1054463874964, 57.1767073742773)}
        \def\eomtmaskedLdata{(60.5816308781363, 56.2357868966477)}
        \def\eomtmaskedBdata{(104.210289458541, 51.503)}
        \def\eomtmaskedSdata{(108.12510940451, 46.0803150765784)}

        \def\eomtgdata{(55.0261783401545, 57.0337931383071)}
        \def\eomtLdata{(127.6, 56.0252744290813)}
        \def\eomtBdata{(260.617881105863, 50.559)}
        \def\eomtSdata{(330.230118412222, 44.6771685888311)}

        \begin{axis}[
            font=\small,
            width=1.06\linewidth,
            height=0.68\linewidth,
            xmode=log,
            xlabel={Frames per second (FPS) [log scale]},
            ylabel={Panoptic Quality (PQ)},
            ylabel style={yshift=-0.5em},
            xlabel style={yshift=0.4em},
            xmin=16, xmax=360,
            ymin=41.5, ymax=58.5,
            xtick={20, 40, 80, 160, 320},
            xticklabels={20, 40, 80, 160, 320},
            ytick={42, 46, 50, 54, 58},
            legend style={font=\scriptsize, at={(0.01,0.01)}, anchor=south west},
            legend cell align=left,
            ymajorgrids=true,
            xmajorgrids=true,
            grid style=dashed,
            tick label style={font=\footnotesize},
            label style={font=\footnotesize},
        ]

        \addlegendimage{empty legend}
        \addlegendentry{\hspace{-11.7pt}%
        \tikz \node[mark size=2pt, inner sep=0.8pt] at (0, 0) {\pgfuseplotmark{\vitgmark}};~~ViT-g~~
        \tikz \node[mark size=2pt, inner sep=0.8pt] at (0, 0) {\pgfuseplotmark{\vitLmark}};~~ViT-L~~
        \tikz \node[mark size=2pt, inner sep=0.8pt] at (0, 0) {\pgfuseplotmark{\vitBmark}};~~ViT-B~~
        \tikz \node[mark size=2pt, inner sep=0.8pt] at (0, 0) {\pgfuseplotmark{\vitSmark}};~~ViT-S
        };
        \addlegendimage{color=adapterlinecolor,
            mark=*,
            line width=0.75pt}
        \addlegendentry{ViT-Adapter + Mask2Former}
        \addlegendimage{color=eomtmaskedcolor,
            mark=*,
            line width=0.75pt,
            densely dashed}
        \addlegendentry{EoMT w/ Masked Attention}
        \addlegendimage{color=eomtcolor,
            mark=*,
            line width=0.75pt}
        \addlegendentry{\textbf{EoMT} (Ours)}

        \tikzset{vitadapter line style/.style={color=adapterlinecolor,line width=0.75pt}}
        \tikzset{eomtmasked line style/.style={color=eomtmaskedcolor,line width=0.75pt}}
        \tikzset{eomt line style/.style={color=eomtcolor,line width=0.75pt}}
]

        \addplot[vitadapter line style, mark=\vitgmark] coordinates {\vitadaptergdata};
        \addplot[vitadapter line style, mark=\vitLmark] coordinates {\vitadapterLdata};
        \addplot[vitadapter line style, mark=\vitBmark] coordinates {\vitadapterBdata};
        \addplot[vitadapter line style, mark=\vitSmark] coordinates {\vitadapterSdata};
        \addplot[vitadapter line style,
            mark=none,
        ]
        coordinates {
            \vitadaptergdata
            \vitadapterLdata
            \vitadapterBdata
            \vitadapterSdata
        };

        \addplot[eomtmasked line style, mark=\vitgmark] coordinates {\eomtmaskedgdata};
        \addplot[eomtmasked line style, mark=\vitLmark] coordinates {\eomtmaskedLdata};
        \addplot[eomtmasked line style, mark=\vitBmark] coordinates {\eomtmaskedBdata};
        \addplot[eomtmasked line style, mark=\vitSmark] coordinates {\eomtmaskedSdata};
        \addplot[eomtmasked line style, densely dashed,
            mark=none,
        ]
        coordinates {
            \eomtmaskedgdata
            \eomtmaskedLdata
            \eomtmaskedBdata
            \eomtmaskedSdata
        };

        \addplot[eomt line style, mark=\vitgmark] coordinates {\eomtgdata};
        \addplot[eomt line style, mark=\vitLmark] coordinates {\eomtLdata};
        \addplot[eomt line style, mark=\vitBmark] coordinates {\eomtBdata};
        \addplot[eomt line style, mark=\vitSmark] coordinates {\eomtSdata};
        \addplot[eomt line style,
            mark=none,
        ]
        coordinates {
            \eomtgdata
            \eomtLdata
            \eomtBdata
            \eomtSdata
        };

        \end{axis}

        \def\blockwidth{1.1cm}
        \def\blockheight{1.8cm}
        \node (vit1) [draw=adapterlinecolor, fill=adapterlinecolor!20!white, thick, inner sep=0pt, draw, anchor=west, rounded corners, minimum width=\blockwidth, minimum height=\blockheight, font=\scriptsize] at (-0.25cm, 5.7cm) {ViT};
        \node (adapter) [draw=adapterlinecolor, fill=adapterlinecolor!20!white, thick,inner sep=0pt, draw, anchor=west, rounded corners, minimum width=\blockwidth, minimum height=\blockheight, font=\scriptsize, right=0.3cm of vit1]  {Adapter};
        \node (pixeldecoder) [draw=adapterlinecolor, fill=adapterlinecolor!20!white, thick,inner sep=0pt, draw, anchor=west, rounded corners, minimum width=\blockwidth, minimum height=\blockheight, text width=1cm, align=center, font=\scriptsize, right=0.3cm of adapter]  {Pixel\\Decoder};
        \node (tfdecoder) [draw=adapterlinecolor, fill=adapterlinecolor!20!white, thick,inner sep=0pt, draw, anchor=west, rounded corners, minimum width=\blockwidth, minimum height=\blockheight, text width=0.8cm, align=center, font=\scriptsize, right=0.3cm of pixeldecoder]  {TF\\Decoder};

        \node (vit2) [draw=eomtcolor, fill=eomtcolor!20!white, thick,inner sep=0pt, draw, anchor=west, rounded corners, minimum width=\blockwidth, minimum height=\blockheight, align=center, font=\scriptsize, right=0.85cm of tfdecoder]  {ViT};

        \node (image1) [inner sep=1pt, anchor=south, align=center, font=\scriptsize] at ($(vit1.north)!0.5!(adapter.north)+(0,0.17cm)$){Image};
        \draw[-stealth] (image1.east) -|(adapter.north);
        \draw[-stealth] (image1.west) -|(vit1.north);

        \draw[-stealth] ($(adapter.west)+(0,0.7cm)$) -- ($(vit1.east)+(0,0.7cm)$);
        \draw[-stealth] ($(vit1.east)+(0,0.6cm)$) -- ($(adapter.west)+(0,0.6cm)$);

        \draw[-stealth] ($(adapter.west)+(0,0.267cm)$) -- ($(vit1.east)+(0,0.267cm)$);
        \draw[-stealth] ($(vit1.east)+(0,0.167cm)$) -- ($(adapter.west)+(0,0.167cm)$);

        \draw[-stealth] ($(adapter.west)+(0,-0.167cm)$) -- ($(vit1.east)+(0,-0.167cm)$);
        \draw[-stealth] ($(vit1.east)+(0,-0.267cm)$) -- ($(adapter.west)+(0,-0.267cm)$);

        \draw[-stealth] ($(adapter.west)+(0,-0.6cm)$) -- ($(vit1.east)+(0,-0.6cm)$);
        \draw[-stealth] ($(vit1.east)+(0,-0.7cm)$) -- ($(adapter.west)+(0,-0.7cm)$);

        \draw[-stealth] ($(adapter.east)+(0,0.65cm)$) -- ($(pixeldecoder.west)+(0,0.65cm)$);

        \draw[-stealth] ($(adapter.east)+(0,0.217cm)$) -- ($(pixeldecoder.west)+(0,0.217cm)$);

        \draw[-stealth] ($(adapter.east)+(0,-0.217cm)$) -- ($(pixeldecoder.west)+(0,-0.217cm)$);

        \draw[-stealth] ($(adapter.east)+(0,-0.65cm)$) -- ($(pixeldecoder.west)+(0,-0.65cm)$);

        \draw[-stealth] ($(pixeldecoder.east)+(0,0.65cm)$) -- ($(tfdecoder.west)+(0,0.65cm)$);

        \draw[-stealth] ($(pixeldecoder.east)+(0,0.217cm)$) -- ($(tfdecoder.west)+(0,0.217cm)$);

        \draw[-stealth] ($(pixeldecoder.east)+(0,-0.217cm)$) -- ($(tfdecoder.west)+(0,-0.217cm)$);
22
        \draw[-stealth] ($(pixeldecoder.east)+(0,-0.65cm)$) -- ($(tfdecoder.west)+(0,-0.65cm)$);

        \node (queries1) [inner sep=1pt, anchor=south, align=center, font=\scriptsize] at ($(tfdecoder.north)+(0,0.17cm)$){Queries};
        \draw[-stealth] (queries1.south) --(tfdecoder.north);

        \node (image2) [inner sep=1pt, anchor=south, align=center, font=\scriptsize] at ($(vit2.north)+(0,0.17cm)$){Image};
        \draw[-stealth] (image2.south) -|(vit2.north);

        \node (queries2) [rotate=90, inner sep=1pt, anchor=south, align=center, font=\scriptsize] at ($(vit2.west)+(-0.17,-0.4cm)$){Queries};
        \draw[-stealth] (queries2.south) --(vit2.west |- queries2.south);

        \node (output1) [inner sep=1pt, anchor=north, align=center, font=\scriptsize] at ($(tfdecoder.south)+(0,-0.19cm)$){~\,Masks~+~Classes};
        \draw[-stealth] (tfdecoder.south) --(output1.north);

        \node (output2) [inner sep=1pt, anchor=north, align=center, font=\scriptsize] at ($(vit2.south)+(0,-0.19cm)$){~\,Masks~+~Classes};
        \draw[-stealth] (vit2.south) --(output2.north);

        \draw[-, dashed, gray] ($(output1.south)!0.47!(output2.south)$) -- ++(0,2.8cm);

        \coordinate (vitadapter) at ($(vit1.north)!0.5!(tfdecoder.north)$);
        \node [align=center, above=0.45cm of vitadapter, anchor=south, font=\footnotesize, text depth=0] {\textbf{ViT-Adapter + Mask2Former~\cite{chen2023vitadapter,cheng2022mask2former}}}; 
        \node [align=center, above=0.45cm of vit2.north, anchor=south, font=\footnotesize, text depth=0] {\textbf{EoMT (Ours)}}; 
        
    \end{tikzpicture} 
    \vspace{-15pt}
    \caption{\textbf{ViT-Adapter + Mask2Former \vs~EoMT (Ours).}  EoMT demonstrates an optimal balance between Panoptic Quality (PQ) and FPS across different sizes of DINOv2~\cite{oquab2023dinov2} pre-trained ViTs~\cite{dosovitskiy2021vit}. Evaluation on COCO \textit{val2017}~\cite{lin2014coco}, see \cref{tab:model_size}.
    }
    \label{fig:comparison}
\end{figure}

The Vision Transformer (ViT)~\cite{dosovitskiy2021vit} has proven to be a strong, scalable, and generally applicable architecture for computer vision~\cite{dehghani2023vit22b, zhai2022scaling, oquab2023dinov2, alabdulmohsin2023sovit}.
Recently, research has shown that ViTs are very suitable for large-scale pre-training~\cite{oquab2023dinov2,zhai2023siglip,fang2024eva02,bao2022beit}, resulting in generalizable models that achieve high performance on many downstream tasks.
A particularly well-researched task is image segmentation, \eg, semantic, instance, and panoptic segmentation~\cite{kirillov2019panoptic}.
To achieve state-of-the-art segmentation performance with ViTs, they are typically combined with several computationally intensive and task-specific components such as ViT-Adapter and Mask2Former (M2F)~\cite{chen2023vitadapter,cheng2022mask2former}.
In these methods, an \textit{adapter}~\cite{chen2023vitadapter,xia2024vitcomer} is first applied in parallel to the ViT, using convolutional layers and interacting with the ViT to extract multi-scale features.
Second, these multi-scale features are fed to a Mask Transformer module~\cite{cheng2022mask2former,cheng2021maskformer,jain2023oneformer, yu2022kmaxdeeplab}, consisting of a \textit{pixel decoder} and a \textit{Transformer decoder}.
The pixel decoder fuses and enhances information across multiple feature scales.
The Transformer decoder then introduces learnable object queries that attend to the multi-scale features via cross-attention. These queries are finally used to generate segmentation mask and class predictions.

In this paper, we explore whether these additional components, as visualized in \cref{fig:comparison}~(top left), are truly necessary to obtain state-of-the-art performance.
We hypothesize that with increasingly extensive pre-training and for larger ViTs, the positive effect of these additional components decreases, making them nearly irrelevant.
There are two key reasons for this:
($i$) Large-scale pre-training, particularly when combined with objectives like masked image modeling (\eg, DINOv2~\cite{oquab2023dinov2}), teaches the ViT to extract dense, fine-grained semantic information essential for segmentation~\cite{kerssies2024benchmark}.
Therefore, we expect that additional components are no longer required to aid the ViT in extracting this information. 
($ii$) Larger ViTs have more learnable parameters. We expect that this increased capacity allows large ViTs to accurately conduct image segmentation without additional components.
If these hypotheses hold, the simplicity and efficiency of segmentation models could be significantly enhanced by removing these task-specific components, while only minimally affecting their segmentation accuracy.

To confirm these hypotheses, we experimentally assess the effect of gradually removing the aforementioned components, in combination with several types of pre-training and different model sizes.
This process ultimately leads us to a conceptually simple model with an architecture that only minimally differs from the plain ViT.
We call this model the \textit{Encoder-only Mask Transformer} (EoMT). 
Crucially, EoMT repurposes the ViT blocks to not only extract image features, but also enable interaction between learnable object queries and image features, to finally predict a mask and class label for each of these queries.
This highly simplified design is visualized in \cref{fig:comparison} (top right). 

A key innovation of EoMT is that it does not require masked attention during inference, unlike existing Mask2Former-like architectures. 
Models use masked attention to constrain each query to cross-attend only within the intermediate segmentation mask predicted for that query.
While this improves segmentation accuracy, it also harms efficiency as it requires additional operations.
To overcome this, we present a novel \textit{mask annealing} strategy, where masked attention is initially fully enabled during training, but then gradually phased out as training progresses, allowing for efficient, masked-attention-free inference.

Overall, EoMT offers several advantages:
($i$) By not requiring additional components and by enabling inference without masked attention, EoMT greatly reduces computational requirements and latency.
($ii$) Due to its architectural simplicity, EoMT is significantly easier to implement than existing approaches that use additional components.
($iii$) By relying purely on the Transformer architecture~\cite{vaswani2017transformer} of the ViT, EoMT can fully and directly leverage ongoing and future developments related to Transformers, without being bottlenecked by additional non-optimized modules.
This applies not only to general improvements like FlashAttention~\cite{dao2022flashattention,dao2024flashattention2} and specialized hardware~\cite{elster2022hopper,choquette2023h100}, but also to vision-specific advances like token merging~\cite{norouzi2024algm,bolya2023tome,lu2023cts} and vision foundation models~\cite{oquab2023dinov2, fang2024eva02}.

Through our experimental analysis, we find that removing task-specific components has only a minimal impact on segmentation accuracy when using a large pre-trained model like DINOv2~\cite{oquab2023dinov2}, confirming our hypotheses. 
By removing these components, EoMT achieves performance competitive with the state of the art, while being much faster. As shown in \cref{fig:comparison}, EoMT obtains a considerably better balance between prediction speed and segmentation quality. 
To illustrate this: ViT-Adapter + M2F with \mbox{ViT-B} achieves a Panoptic Quality (PQ) of 54.4 at 32 frames per second (FPS). In contrast, despite using a larger model, EoMT runs significantly faster with \mbox{ViT-L} at 128~FPS while also achieving a higher PQ of 56.0.

In summary, we make the following contributions:
    \textbf{(1)} We assess the necessity of task-specific components of state-of-the-art image segmentation models and find that they become less relevant when scaling up model size and pre-training.
    \textbf{(2)} We present the \textit{Encoder-only Mask Transformer} (EoMT), a simple and efficient model that repurposes the ViT blocks for segmentation and obtains state-of-the-art performance without requiring inefficient task-specific components.
    \textbf{(3)} We propose a \textit{mask annealing} training strategy that enables significantly faster inference by removing the need for masked attention.

\section{Related Work}
\label{sec:related}

\PARbegin{Image segmentation.}
Image segmentation is a fundamental task in computer vision, for which the goal is to divide an image into pixel-level segments based on semantics, by providing a segmentation mask and class label for each segment. 
For semantic segmentation, the objective is to output a single segment for each class in the image. Instance segmentation, on the other hand, requires a segment for each individual object instance, but disregards uncountable entities like `road' or `sky'. Finally, panoptic segmentation~\cite{kirillov2019panoptic} combines these tasks and requires ($i$) segments per individual instance for countable
classes called \textit{things} (\eg, `person' or `car'), and ($ii$) a single segment per class for uncountable 
classes called \textit{stuff} (\eg, `road' or `sky').

Traditionally, segmentation methods had specialized architectures that were tailored for only one of these tasks~\cite{zhao2017pspnet,chen2018deeplabv3plus,kirillov2019panopticfpn,he2017maskrcnn,strudel2021segmenter}. Recently, however, the emergence of the \textit{Mask Transformer} framework~\cite{carion2020end, wang2021max,cheng2021maskformer} has enabled the adoption of a unified architecture and training pipeline for all three segmentation tasks~\cite{cheng2021maskformer,cheng2022mask2former,yu2022kmaxdeeplab,cavagnero2024pem,jain2023oneformer,li2023maskdino,athar2023tarvis,OMGSeg}. The versatility of these models is enabled by learnable object queries, which adaptively learn to represent a single segment, whether it is a \textit{stuff} class or a \textit{thing} instance. In this work, we investigate state-of-the-art Mask Transformer models and propose a minimalistic, efficient ViT-based model that is competitive with more complex architectures while being significantly more efficient.

\PAR{Vision Transformers.}
The Transformer architecture~\cite{vaswani2017transformer} was originally developed for Natural Language Processing (NLP), where its ability to model long-range dependencies revolutionized the field. To harness the power of this architecture for computer vision, the Vision Transformer (ViT)~\cite{dosovitskiy2021vit} was introduced. ViTs divide images into fixed-size patches, project them into an embedding space to form tokens, and then process these tokens using multiple Transformer blocks. By design, there are key differences between Convolutional Neural Networks (CNNs) and ViTs. ($i$) ViTs process images at a fixed resolution due to the fixed patch size. In contrast, CNNs (\eg, ResNet~\cite{he2016resnet}) typically contain various downscaling steps, allowing them to output feature maps of multiple resolutions~\cite{lin2017fpn}. ($ii$) ViTs process images globally leveraging self-attention, while CNNs process images locally by applying convolutional filters. 

For tasks like image segmentation, multi-scale features and local processing are claimed to be beneficial for performance~\cite{chen2023vitadapter}. To introduce these properties into Transformers, some works propose alternative architectures~\cite{liu2021swin,hassani2023nat,hassani2022dinat,xie2021segformer} that incorporate local attention and token downsampling. However, as discussed later, these models deviate from the plain ViT architecture, preventing them from leveraging advancements in large-scale pre-training. An alternative approach extends ViTs with a CNN-based \textit{adapter} to produce multi-scale features~\cite{chen2023vitadapter,xia2024vitcomer}. In this work, however, we find that the necessity of these adapters and other task-specific components greatly diminishes when using extensively pre-trained large ViTs for image segmentation. This allows a much simpler and more efficient model while maintaining performance competitive with state-of-the-art approaches. Similar to our work, UViT~\cite{chen2022simple} explores the use of single-scale features from the plain ViT for instance recognition tasks, but it still relies on complex task-specific decoders. Meanwhile, YOLOS~\cite{fang2021you} adopts an encoder-only ViT for instance recognition but is restricted exclusively to object detection. Moreover, neither method demonstrates that scaling up model size and pre-training enables a simple ViT-based model to be competitive with complex state-of-the-art architectures, which we do with EoMT.

\PAR{Large-scale visual pre-training.}
For tasks like image segmentation, it is common practice to initialize a model's backbone with pre-trained weights to improve downstream performance over random initialization. Initially, such pre-training relied on supervised image classification on ImageNet~\cite{deng2009imagenet}. More recently, pre-training has been scaled up to massive datasets using weakly-~\cite{radford2021clip} or self-supervised~\cite{caron2021dino} learning. Currently, vision foundation models (VFMs) that use masked image modeling, like DINOv2~\cite{oquab2023dinov2} and EVA-02~\cite{fang2024eva02}, provide the best pre-training for image segmentation~\cite{kerssies2024benchmark}. Notably, all these VFMs use the ViT architecture for its scalability. This means that models with non-ViT backbones, such as Swin~\cite{liu2021swin} and ConvNeXt~\cite{liu2022convnext}, which are commonly used for image segmentation, cannot leverage VFM pre-training due to their incompatible architectures.
In contrast, EoMT \textit{can} benefit from VFM initialization, as it is fully ViT-based.

\section{Towards Encoder-only Mask Transformer}
\label{sec:method}

\subsection{Preliminaries}

\PARbegin{Vision Transformers.}
Vision Transformers first divide an image $\bm{I} \in \mathbb{R}^{3\times{H}\times{W}}$ into $N$ non-overlapping patches of shape $({p}\times{p})$, where $H$ and $W$ are the image height and width, and $p$ is the pre-determined patch size. Subsequently, these patches are linearly projected into patch tokens $\bm{X}^0 \in \mathbb{R}^{{D}\times{N}}$ and processed by $L$ Transformer blocks~\cite{vaswani2017transformer}. Each Transformer block applies multi-head self-attention (\texttt{MHSA}) and a two-layer multi-layer perceptron (\texttt{MLP}) with a non-linear activation function. Concretely, for each block $i$,
\begin{equation}
\begin{split}
    \bm{Z}^i & = \bm{X}^i + \texttt{MHSA}(\texttt{Norm}(\bm{X}^i)); \label{eq:vit} \\
    \bm{X}^{i+1} & = \bm{Z}^i + \texttt{MLP}(\texttt{Norm}(\bm{Z}^i)),
\end{split}
\end{equation}
where \texttt{Norm} is Layer Normalization~\cite{ba2016layernorm}. The result of this process is a set of final patch tokens $\bm{X}^L$. Reordering these tokens yields spatial features $\bm{F}^\textrm{vit} \in \mathbb{R}^{{D}\times{\frac{H}{p}}\times{\frac{W}{p}}}$, where the patch size $p$ determines the resolution.

To achieve state-of-the-art image segmentation with ViTs, recent works have proposed several components that further process the resulting patch tokens and interact with the ViT at different levels.

\PAR{Adapters.}
To introduce convolutional biases and enable multi-scale feature extraction, an \textit{adapter} is applied in parallel to the ViT to inject and extract features~\cite{chen2023vitadapter,xia2024vitcomer}. Concretely, the ViT-Adapter~\cite{chen2023vitadapter}, which we study in this work, first applies a CNN to the input image to extract multi-scale features at resolutions $\frac{1}{4}$, $\frac{1}{8}$, $\frac{1}{16}$, and $\frac{1}{32}$. Subsequently, the ViT-Adapter repeatedly injects these CNN features into the ViT through multi-scale deformable attention~\cite{zhu2021deformabledetr}, applies several ViT blocks, and extracts refined features from the ViT into the multi-scale CNN features. The output of the adapter is a set of multi-scale ViT- and CNN-based features, $\{\bm{F}_{4}, \bm{F}_{8}, \bm{F}_{16}, \bm{F}_{32}\}$, with $\bm{F}_{i} \in \mathbb{R}^{D \times \frac{H}{i} \times \frac{W}{i}}$.

\PAR{Mask Transformers.}
To make segmentation predictions with these features, state-of-the-art models follow the \textit{Mask Transformer} framework~\cite{carion2020end, wang2021max, cheng2021maskformer}.
In this work, we study the state-of-the-art method Mask2Former (M2F)~\cite{cheng2022mask2former}.
As a first step, to further enhance the features extracted by the ViT-Adapter, M2F applies a \textit{pixel decoder}.
This pixel decoder takes the features $\{ \bm{F}_{4}, \bm{F}_{8}, \bm{F}_{16}, \bm{F}_{32}\}$ and applies a series of multi-scale deformable attention layers~\cite{zhu2021deformabledetr}, outputting processed features $\{ \bm{\widehat{F}}_{4}, \bm{\widehat{F}}_{8}, \bm{\widehat{F}}_{16}, \bm{\widehat{F}}_{32}\}$.
In this process, multi-scale features from different backbone layers are processed into a consistent yet scale-specific representation.

The final component of M2F, the \textit{Transformer decoder}, generates and outputs the actual segmentation predictions.
As inputs, it takes not only the multi-scale features from the pixel decoder, but also a set of $K$ learned queries $\mathcal{Q}^0 = \{\bm{q}^0_i \in \mathbb{R}^{D}\}^{K}_{i=1}$. In the Transformer decoder, each of these queries learns to represent one individual segment per image (\ie, a \textit{stuff} class or a \textit{thing} instance).
To do so, these queries are subjected to $J$ blocks with cross-attention to the multi-scale features and multi-head self-attention between queries, yielding processed queries ${\mathcal{Q}^{J}}$. 
The Transformer decoder then generates the segmentation predictions by predicting a class label and segmentation mask for each query $\bm{q}^J_i$. Class logits $\bm{c}^J_i \in \mathbb{R}^{C}$ are predicted by applying a linear layer to $\bm{q}^J_i$. Mask logits $\bm{m}^J_{i} \in \mathbb{R}^{\frac{H}{4}\times\frac{W}{4}}$ are obtained by first applying an \texttt{MLP} to yield mask embedding $\bm{\widehat{q}}^J_i$ and subsequently taking the dot product between the features ${\bm{\widehat{F}}}_{4}$ and $\bm{\widehat{q}}^J_i$. By assigning each ground-truth segment to a unique query and supervising both class and mask predictions, the model learns the segmentation task.

A key feature of the M2F Transformer decoder is the use of masked cross-attention.
In each of the $J$ blocks, prior to cross-attention between queries and image features, an intermediate segmentation mask and class label are predicted and supervised per query using the above procedure. %
The predicted masks are then used to mask the attention, allowing a query to only attend to the image region that corresponds to its predicted segmentation mask.
This masked attention results in improved segmentation accuracy~\cite{cheng2022mask2former}.

\begin{figure}[t]
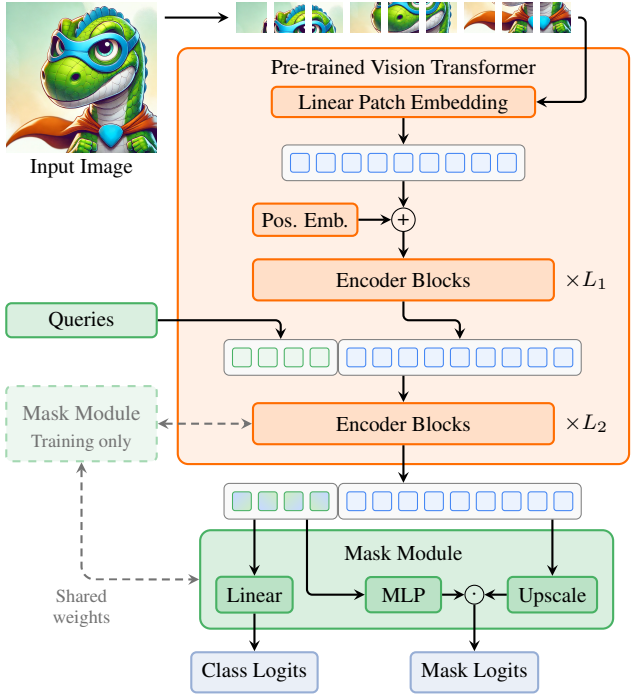
%
    \centering%
    \include{figures/arch}%
    \vspace{-0.7cm}
    \caption{\textbf{EoMT architecture.} Learnable queries are concatenated to the patch tokens after the first $L_1$ ViT encoder blocks. These concatenated tokens are then jointly processed by the last $L_2$ blocks and used to predict class and mask logits. 
    }
    \label{fig:arch}
\end{figure}

\subsection{Removing Task-Specific Components}
\label{sec:method:steps}
To study the importance of the aforementioned task-specific components, we gradually remove them while assessing the effect on segmentation accuracy.
We continue until all task-specific components are removed, ending up with our simple \textit{Encoder-only Mask Transformer} (EoMT). The different configurations are described in this section and visualized in \cref{fig:supp:steps}, with the results reported in \cref{sec:exp:main}.

\PAR{Removing the adapter.}
Removing the adapter eliminates the inductive biases and multi-resolution features provided by a CNN. However, we hypothesize that with sufficient pre-training, large ViTs can learn these features without requiring  additional components.
In the absence of an adapter, we construct a feature pyramid by upscaling the ViT output features $\bm{F}^\textrm{vit}=\bm{F}_{16}$ (patch size $16\times16$) to compute ${\bm{F}_{4}}$ and ${\bm{F}_{8}}$, and downscaling them to compute $\bm{F}_{32}$.
We follow the approach of ViTDet~\cite{li2022vitdet} by using transposed convolutions for upscaling and normal convolutions for downscaling.
For each scale of the feature pyramid, we independently up- or downscale $\bm{F}^\textrm{vit}$ with a sequence of operations.
We repeat a (transposed) $2\times2$ convolution with stride $2\times2$, GELU activation, depthwise $3\times3$ convolution, and final \texttt{Norm}, until the required scales are reached.
This approach mimics the multi-scale feature extraction of the ViT-Adapter in a much simpler manner.

\PAR{Removing the pixel decoder.} Without the adapter, the resulting features no longer originate from different stages of a hierarchical backbone and thus should not need further consolidation. As such, the heavy pixel decoder should be obsolete, and we remove it by directly feeding the simplified feature pyramid to the Transformer decoder.
Specifically, instead of $\{ \bm{\widehat{F}}_{4}, \bm{\widehat{F}}_{8}, \bm{\widehat{F}}_{16}, \bm{\widehat{F}}_{32}\}$, we input $\{\bm{F}_{4}, \bm{F}_{8}, \bm{F}_{16}, \bm{F}_{32}\}$ to the Transformer decoder.

\PAR{Removing multi-scale feature processing.}
To further simplify, we question the necessity of using features at multiple scales, since all features are derived from a shared single-scale feature map $\bm{F}^\textrm{vit}$. Therefore, we do not generate multi-scale features ${\bm{F}_{8}, \bm{F}_{32}}$. In the Transformer decoder, instead, queries cross-attend exclusively to the ViT output $\bm{F}^\textrm{vit}=\bm{F}_{16}$. We only upscale $\bm{F}^\textrm{vit}$ to $\bm{F}_{4}$ to compute the mask logits via dot product with the mask embeddings $\bm{\widehat{q}}^J_i$, to ensure high-resolution output masks.

\subsection{Encoder-only Mask Transformer}

By removing the previous task-specific components, the model is reduced to a ViT with a single-scale Transformer decoder. The next step is to completely remove the decoder. This requires minor modifications to the plain ViT architecture, such that it can perform image segmentation without a dedicated decoder. We call the resulting method the \textit{Encoder-only Mask Transformer} (EoMT).

\PAR{Querying the ViT for masks.}
Differently from standard Mask Transformers for image segmentation~\cite{cheng2021maskformer, cheng2022mask2former, jain2023oneformer, li2023maskdino, yu2022kmaxdeeplab, hu2023yoso}, which adopt heavy and complex ad-hoc decoders, EoMT only uses the architecture of the plain ViT with a few extra learned queries and a small mask prediction module. An overview of EoMT is provided in \cref{fig:arch}.

The first $L_1$ Transformer encoder blocks of the ViT are unchanged and only process the input image.
After these encoder blocks, we introduce a set of $K$ learnable queries that are concatenated to the patch tokens.
These are then jointly processed by the remaining $L_2$ ViT encoder blocks, following \cref{eq:vit}. 
The final blocks thus have to process the patch tokens as before, but also replace the Transformer decoder that processes the queries.

\begin{figure}[t]
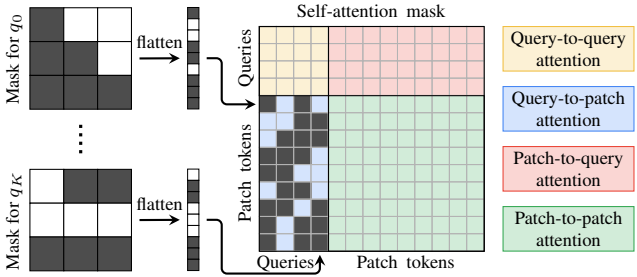

    \centering
    \include{figures/attention_masking}%
    \vspace{-24pt}
    \caption{\textbf{Masked self-attention during training.} In the final $L_2$ blocks of EoMT, patch tokens and queries are jointly processed by self-attention. During training, the intermediate mask predictions are used to mask the query-to-patch portion of the attention operation, mimicking the masked cross-attention of M2F~\cite{cheng2022mask2former}.}
    \label{fig:masked_self_attention}
\end{figure}

A standard Mask Transformer introduces ($i$) interaction between individual queries through self-attention, enabling queries to coordinate the objects they should attend to, and ($ii$) transfer of information from visual tokens to object queries through query-to-patch cross-attention. Normally, these operations are performed sequentially. In contrast, by using the \texttt{MHSA} operation of the ViT, EoMT performs them jointly in a single layer (see Fig.~\ref{fig:masked_self_attention}).

In addition to the ViT, we introduce a small \textit{mask module} to predict masks and corresponding classes for each query.
Here, we follow the same design as M2F~\cite{cheng2022mask2former}, passing the query tokens through a linear layer to predict class logits and using a three-layer \texttt{MLP} followed by a dot product with the upscaled image features $\bm{F}_4$ to obtain mask logits.

To enable masked attention between queries and image features during training, as in the M2F Transformer decoder, we additionally apply the previously introduced mask module before each of the last $L_2$ ViT blocks, to predict intermediate segmentation masks for each query.
In turn, these masks can be used to mask the query-to-patch attention in the plain self-attention block, constraining the attention of each query to the segmentation mask that is predicted for it. 
This is visualized in \cref{fig:masked_self_attention}.

\begin{figure}[t]
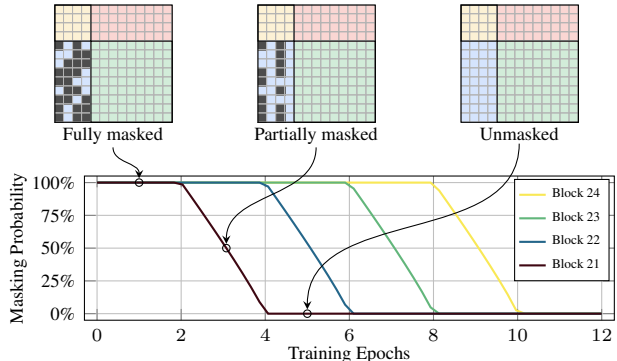

    \centering
    \include{figures/mask_annealing}%
    \vspace{-32pt}
    \caption{\textbf{Mask annealing during training.}
    Self-attention is initially masked in the final $L_2$ ($=4$ for ViT-L) EoMT blocks. The masking probability is gradually annealed, starting from early blocks, until it is no longer needed at the end of training.
    }
    \label{fig:mask_annealing}
\end{figure}

\PAR{Mask annealing.}
While using masked attention during training improves performance, predicting intermediate masks and applying them to the self-attention operation for each block during inference is computationally expensive and inefficient.
To address this, we propose a \textit{mask annealing} scheme that gradually phases out masked attention over the course of training.
Specifically, in each block with masked attention, we apply the mask for each query with a probability $P_\textrm{mask}$, which starts at 1.0 and decays block by block to 0.0 throughout the training, as shown in~\cref{fig:mask_annealing}.
This strategy allows the model to initially benefit from masked attention to aid convergence, while gradually learning to operate without it, thus maintaining high performance. By eliminating the need for masked attention during inference, the final model leverages the highly optimized plain ViT architecture without additional intermediate modules or modified attention operations, ensuring optimal efficiency.

\section{Experiments}
\label{sec:exp}

\subsection{Experimental Setup}

\PAR{Datasets.} We use widely adopted benchmarks for image segmentation: COCO~\cite{lin2014coco} and ADE20K~\cite{zhou2017ade20k} for panoptic, Cityscapes~\cite{cordts2016cityscapes} and ADE20K for semantic, and COCO for instance segmentation.

\PAR{Models and training.}
Unless specified otherwise, we use DINOv2-L~\cite{oquab2023dinov2} with a $640\times640$ input size and a $16\times16$ patch size. For EoMT with \mbox{ViT-L}, we use $L_{2} = 4$, which provides optimal performance (see \cref{sec:supp:exp} and \cref{tab:supp:blocks}).
We train our models in mixed precision with the AdamW~\cite{loshchilov2019adamw} optimizer (learning rate $10^{-4}$), layer-wise learning rate decay (LLRD)~\cite{bao2022beit} (factor 0.8), polynomial learning rate decay (factor 0.9), and polynomial mask annealing (factor 0.9). The number of epochs is set to 12 ($1\times$ schedule~\cite{wu2019detectron2}) on COCO, 31~\cite{kerssies2024benchmark} on ADE20K, and 107~\cite{kerssies2024benchmark} on Cityscapes.

\PAR{Evaluation.}
We use standard evaluation metrics: Panoptic Quality (PQ) for panoptic~\cite{kirillov2019panoptic}, mean Intersection over Union (mIoU) for semantic~\cite{everingham2010pascal}, and Average Precision (AP) for instance segmentation~\cite{lin2014coco}.
We evaluate model efficiency in terms of average inference speed in \textit{frames per second} (FPS) and average number of \textit{floating point operations} (FLOPs) over all images in the validation set, as well as the number of model parameters.
We use an NVIDIA H100 GPU, FlashAttention-2~\cite{dao2024flashattention2}, \texttt{torch.compile}~\cite{ansel2024pytorch2}, and a batch size of 1, unless otherwise specified. The FLOPs are obtained using \textit{fvcore}~\cite{meta2023fvcore} and reported in terms of GFLOPs ($\text{FLOPs} \times 10^9$).

\vskip3pt \noindent 
See \cref{sec:supp:impl_details}
for additional implementation details.

\subsection{Main Results}
\label{sec:exp:main}

\begin{table}
    \centering
    \footnotesize
    \setlength{\tabcolsep}{1.0pt}

    \begin{tabularx}{\linewidth}{
        c
        lYY@{\hskip -3pt}c@{\hskip -5pt}c
        }
        \toprule
        
        & 
        Method & Params & GFLOPs & FPS & PQ \\
        
        \midrule

        \tablestep{0} & 
        ViT-Adapter + Mask2Former &
        349M &
        830 & 
        29 &
        57.1 \\

        \midrule

        \tablestep{1} &
        \downrightarrow w/o ViT-Adapter & 
        342M & 
        700 & 
        \phantom{\qincnew{6}}36\qincnew{7}
        &
        \phantom{\qdecnew{0.0}}56.7\qdecnew{0.4} \\

        \tablestep{2} &
        \hspace{4pt}\downrightarrow w/o Pixel decoder & 
        337M & 
        685 & 
        \phantom{\qincnew{26}}61\qincnew{25}
        &
        \phantom{\qdecnew{0.0}}56.9\qincnew{0.2} \\

        \tablestep{3} & 
        \hspace{8pt}\downrightarrow  w/o Multi-scale & 
        328M & 
        673 & 
        \phantom{\qincnew{2}}64\qincnew{3}
        &
        \phantom{\qdecnew{0.0}}56.7\qdecnew{0.2} \\

        \tablestep{4} & 
        \hspace{12pt}\downrightarrow w/o Transformer decoder & 
        316M & 
        828 & 
        \phantom{\qincnew{3}}61\qdecnew{3}
        &
        \phantom{\qdecnew{0.0}}56.2\qdecnew{0.5} \\

        \tablestep{5} & 
        \hspace{16pt}\downrightarrow w/o Masking = \textbf{EoMT} &
        316M & 
        669 & 
        \phantom{\qincnew{67}}128\qincnew{67}
        &
        \phantom{\qdecnew{0.0}}56.0\qdecnew{0.2} \\
        \bottomrule

    \end{tabularx}

\caption{\textbf{From ViT-Adapter + Mask2Former to EoMT.} Evaluated on COCO \textit{val2017}~\cite{lin2014coco}.} %
\label{tab:steps}

\end{table}

\PARbegin{From ViT-Adapter + Mask2Former to EoMT.}
In \cref{tab:steps}, we evaluate the stepwise removal of task-specific components from ViT-Adapter + Mask2Former (M2F)~\cite{chen2023vitadapter,cheng2022mask2former} to obtain our proposed EoMT model. We find that removing all task-specific components reduces the PQ only slightly, from 57.1 to 56.0, but increases the prediction speed by a substantial $4.4\times$.
Interestingly, this FPS improvement is much larger than the FLOPs improvement. This is because EoMT relies solely on the plain ViT, allowing it to leverage the highly optimized Transformer architecture without being bottlenecked by custom components that contribute little to the segmentation accuracy.
Looking at the intermediate steps, we find that removing the ViT-Adapter, pixel decoder, and multi-scale features in steps \tablestep{1–3} reduce performance by just 0.4~PQ while making the model $2.2\times$ faster.
Step \tablestep{4} temporarily increases FLOPs and reduces FPS, as the mask module is repeatedly applied to generate intermediate segmentation masks for masked attention, with upscaling being the most compute-intensive part. 
However, in step \tablestep{5}, we remove masked attention entirely, eliminating this overhead. The proposed mask annealing strategy enables masked-attention-free inference, further accelerating the model by $2.1\times$ with minimal impact on the PQ. More detailed results on these steps are provided in
\cref{sec:supp:exp} and \cref{tab:supp:steps_detail}.
Overall, the stepwise removal of task-specific components ultimately yields a model that is significantly faster, remarkably simpler, and nearly as accurate.

\PAR{Impact of pre-training.}
Next, we explore the impact of pre-training.
Specifically, we consider large-scale self- and weakly-supervised pre-training with DINOv2~\cite{oquab2023dinov2} and EVA-02~\cite{fang2024eva02}, respectively, as well as supervised ImageNet-21K and ImageNet-1K pre-training with DeiT III~\cite{touvron2022deit}, all with ViT-L.
The results in \cref{tab:pretraining} show that large-scale pre-training allows EoMT to obtain a similar PQ to ViT-Adapter + M2F. For DINOv2 and EVA-02, the overall PQ gap is only 1.1 or 1.2, whereas it increases to 3.9 and 6.1 for ImageNet-21K~\cite{deng2009imagenet} and ImageNet-1K~\cite{russakovsky2015imagenet}, respectively. 
These results confirm our hypothesis that large-scale pre-training renders complex task-specific components increasingly redundant. %
We expect that the \textit{masked image modeling} pre-training objectives of DINOv2 and EVA-02 contribute to this effect, as it enhances the semantic understanding of patches, which is essential for image segmentation~\cite{kerssies2024benchmark}.

\begin{table}[t]
\centering
\notscriptsize
\setlength{\tabcolsep}{2.0pt}
\begin{tabularx}{\linewidth}{
lcYYY@{\hskip -3pt}Y@{\hskip -3pt}c
}
\toprule

Model && 
Pre-train & 
Params &
GFLOPs & 
FPS & 
PQ \\

\midrule

ViT-Adapter + M2F && 
 &
349M &
830 &
29 &
\phantom{\qdecnew{0.0}}57.1\phantom{\qdecnew{0.0}}  
\\

EoMT w/ Masking && 
DINOv2 & 
316M &
828 &
\phantom{\qincnew{32}}61\qincnew{32} &
\phantom{\qdecnew{0.0}}56.2\qdecnew{0.9}
\\

\textbf{EoMT} &&
 &
316M &
669 &
\phantom{\qdecnew{67}}128\qincnew{99} &
\phantom{\qdecnew{0.0}}56.0\qdecnew{1.1}
\\

\midrule

ViT-Adapter + M2F && 
 &
349M &
829 &
25 &
\phantom{\qdecnew{0.0}}56.7\phantom{\qdecnew{0.0}}
\\

EoMT w/ Masking &&
EVA-02 & 
316M &
826 &
\phantom{\qincnew{27}}52\qincnew{27} &
\phantom{\qdecnew{0.0}}56.0\qdecnew{0.7} \\

\textbf{EoMT} &&
& 
316M &
667 &
\phantom{\qincnew{25}}77\qincnew{52} &
\phantom{\qdecnew{0.0}}55.5\qdecnew{1.2} \\

\midrule

ViT-Adapter + M2F && 
& 
349M &
830 &
29 &
\phantom{\qdecnew{0.0}}53.9\phantom{\qdecnew{0.0}} \\

EoMT w/ Masking && 
IN21K & 
316M &
828 &
\phantom{\qincnew{32}}61\qincnew{32} &
\phantom{\qdecnew{0.0}}51.0\qdecnew{2.9} \\

\textbf{EoMT} &&
& 
316M &
669 &
\phantom{\qincnew{67}}128\qincnew{99} &
\phantom{\qdecnew{0.0}}50.0\qdecnew{3.9} \\

\midrule

ViT-Adapter + M2F &&
& 
349M &
830 &
29 &
\phantom{\qdecnew{0.0}}50.4\phantom{\qdecnew{0.0}} \\

EoMT w/ Masking &&
IN1K & 
316M &
828 &
\phantom{\qincnew{32}}61\qincnew{32} &
\phantom{\qdecnew{0.0}}45.9\qdecnew{4.5} \\

\textbf{EoMT} &&
& 
316M & 
669 &
\phantom{\qincnew{67}}128\qincnew{99} &
\phantom{\qdecnew{0.0}}44.3\qdecnew{6.1} \\

\bottomrule

\end{tabularx}
\caption{\textbf{Pre-training.} EoMT performs significantly better with advanced pre-training, \ie, DINOv2~\cite{oquab2023dinov2} or EVA-02~\cite{fang2024eva02}. Evaluated on COCO \textit{val2017}~\cite{lin2014coco}. %
}
\label{tab:pretraining}
\end{table}

\begin{table}[t]
\centering
\notscriptsize
\setlength{\tabcolsep}{1.0pt}
\begin{tabularx}{\linewidth}{
lYYY@{\hskip -3pt}Y@{\hskip -3pt}c
}
\toprule

Model & 
Size & 
Params &
GFLOPS &
FPS &
PQ \\

\midrule

ViT-Adapter + M2F & 
& 
1209M & 
2510 &
20 &        
\phantom{\qdecnew{0.0}}57.7\phantom{\qdecnew{0.0}}  \\

EoMT w/ Masking & 
g & 
1164M &
2689
&
\phantom{\qdecnew{15}}35\qincnew{15}
&
\phantom{\qdecnew{0.0}}57.2\qdecnew{0.5} \\    

\textbf{EoMT} & 
& 
1164M &
2261
&
\phantom{\qdecnew{35}}55\qincnew{35}
&
\phantom{\qdecnew{0.0}}57.0\qdecnew{0.7} \\

\midrule

ViT-Adapter + M2F & 
&
349M &
830 &
29 &
\phantom{\qdecnew{0.0}}57.1\phantom{\qdecnew{0.0}} \\

EoMT w/ Masking & 
L & 
316M &
828 &
\phantom{\qdecnew{32}}61\qincnew{32} &
\phantom{\qdecnew{0.0}}56.2\qdecnew{0.9} \\

\textbf{EoMT} & 
& 
316M &
669 &
\phantom{\qdecnew{99}}128\qincnew{99} &
\phantom{\qdecnew{0.0}}56.0\qdecnew{1.1} \\

\midrule

ViT-Adapter + M2F & 
& 
121M &
347 &
32 &    
\phantom{\qdecnew{0.0}}54.4\phantom{\qdecnew{0.0}} \\

EoMT w/ Masking & 
B & 
93M &
286
&
\phantom{\qdecnew{72}}104\qincnew{72}
&
\phantom{\qdecnew{0.0}}51.5\qdecnew{2.9} \\

\textbf{EoMT} & 
&
93M &
216
&
\phantom{\qdecnew{229}}261\qincnew{229}
&
\phantom{\qdecnew{0.0}}50.6\qdecnew{3.8} \\

\midrule

ViT-Adapter + M2F & 
&
47M &
165 &
33 &
\phantom{\qdecnew{0.0}}50.5\phantom{\qdecnew{0.0}} \\

EoMT w/ Masking & 
S & 
24M &
89
&
\phantom{\qdecnew{75}}108\qincnew{75}
&
\phantom{\qdecnew{0.0}}46.1\qdecnew{4.4} \\

\textbf{EoMT} & 
& 
24M &
68
&
\phantom{\qdecnew{297}}330\qincnew{297}
&
\phantom{\qdecnew{0.0}}44.7\qdecnew{5.8} \\

\bottomrule

\end{tabularx}
\caption{\textbf{Model size.} EoMT performs significantly better as the ViT~\cite{dosovitskiy2021vit} model size increases. Evaluated on COCO \textit{val2017}~\cite{lin2014coco}.
} 
\label{tab:model_size}
\end{table}

\begin{table*}[t]
    \centering
    \scriptsize
    \renewcommand{\tabcolsep}{3.3pt}
    \begin{tabularx}{\linewidth}
    {llcc l YYYY l YYYY}
        \toprule
        
        \multirow{2}[2]{*}{Method} &
        \multirow{2}[2]{*}{Backbone} &
        \multirow{2}[2]{*}{Pre-training} &
        \multirow{2}[2]{*}{Params} &&
        \multicolumn{4}{c}{COCO \emph{val2017}~\cite{lin2014coco}} && 
        \multicolumn{4}{c}{ADE20K \emph{val}~\cite{zhou2017ade20k}} \\
        \cmidrule{6-9}
        \cmidrule{11-14}
        &
        &
        &
        &&
        Input size & 
        GFLOPs & FPS & PQ &&
        Input size & 
        GFLOPs & FPS & PQ \\

        \midrule

        Mask2Former$^\dagger$~\cite{cheng2022mask2former} & 
        Swin-L~\cite{liu2021swin} &
        IN21K &  
        216M &&
        $800^2$ & 868 & 24 & 57.8 &&
        $640^2$ & -- & 33 & 48.1 \\
        
        kMaX-DeepLab~\cite{yu2022kmaxdeeplab}  & 
        ConvNext-L~\cite{liu2022convnext} & 
        IN21K & 
        232M &&
        $1281^2$ & -- & -- & 58.0 &&
        $1281^2$ & 1302  & -- & 50.9  \\
        
        OneFormer$^\dagger$~\cite{jain2023oneformer} & 
        DiNAT-L~\cite{hassani2022dinat} & 
        IN21K & 
        223M &&
        $800^2$ & 736 & 20 & 58.0 &&
        $1280^2$ & 1369 & 10 & 51.5  \\

        OneFormer$^\dagger$~\cite{jain2023oneformer} & 
        DiNAT-L~\cite{hassani2022dinat} & 
        IN21K & 
        223M &&
        -- &  -- & -- & -- &&
        $1280^2$ & 1369 & 10 & \phantom{\textsuperscript{c}}53.5\textsuperscript{c}  \\
        
        MaskDINO$^\dagger$~\cite{li2023maskdino} & 
        Swin-L~\cite{liu2021swin} & 
        IN21K & 
        223M &&
        $800^2$ & 1326 & 14 & 58.3 &&
        -- & -- & -- & -- \\
        
        \midrule
        
        Mask2Former$^\ddagger$~\cite{cheng2022mask2former} & 
        ViT-Adapter-L$^\ddagger$~\cite{chen2023vitadapter} & 
        DINOv2 & 
        349M & &
        $640^2$ & 830 & 29 & 57.1 &&
        $640^2$ & 830 & 29 & \phantom{\textsuperscript{c}}51.8\textsuperscript{c}  \\

        Mask2Former$^\ddagger$~\cite{cheng2022mask2former} & 
        ViT-Adapter-L$^\ddagger$~\cite{chen2023vitadapter} & 
        DINOv2 & 
        354M &&
        $1280^2$ & 4817 & 10 & 59.7 &&
        $1280^2$ & 4817 & 10 & \phantom{\textsuperscript{c}}53.0\textsuperscript{c} \\
        
        \midrule

        Mask2Former$^\ddagger$~\cite{cheng2022mask2former} & 
        ViT-Adapter-g$^\ddagger$~\cite{chen2023vitadapter} & 
        DINOv2 & 
        1209M &&
        $640^2$ & 2510 & 20 & 57.7 &&
        $640^2$ & 2510 & 20 & \phantom{\textsuperscript{c}}52.6\textsuperscript{c} \\

        Mask2Former$^\ddagger$~\cite{cheng2022mask2former} & 
        ViT-Adapter-g$^\ddagger$~\cite{chen2023vitadapter} & 
        DINOv2 & 
        1216M &&
        $1280^2$ & 13790 & 6 & 59.9 &&
        $1280^2$ & 13790 & 6 & \phantom{\textsuperscript{c}}54.2\textsuperscript{c} \\

        \midrule
        
        \textbf{EoMT} (Ours) & 
        ViT-L~\cite{dosovitskiy2021vit} & 
        DINOv2 & 
        316M &&
        $640^2$ & 669 & 128 & 56.0 &&
        $640^2$ & 669 & 128 & \phantom{\textsuperscript{c}}50.6\textsuperscript{c} \\
        
        \textbf{EoMT} (Ours) & 
        ViT-L~\cite{dosovitskiy2021vit} & 
        DINOv2 & 
        322M &&
        $1280^2$ & 4146 & 30 & 58.3 &&
        $1280^2$ & 4146 & 30 & \phantom{\textsuperscript{c}}51.7\textsuperscript{c} \\

        \midrule

        \textbf{EoMT} (Ours) & 
        ViT-g~\cite{oquab2023dinov2} & 
        DINOv2 & 
        1164M &&
        $640^2$ & 2261 & 55 & 57.0 &&
        $640^2$ & 2261 & 55 & \phantom{\textsuperscript{c}}51.3\textsuperscript{c} \\

        \textbf{EoMT} (Ours) & 
        ViT-g~\cite{oquab2023dinov2} & 
        DINOv2 & 
        1171M &&
        $1280^2$ & 12712 & 12 & 59.2 &&
        $1280^2$ & 12712 & 12 & \phantom{\textsuperscript{c}}52.8\textsuperscript{c} \\

        \bottomrule
    \end{tabularx}
    \caption{\textbf{EoMT for panoptic segmentation.} $^\dagger$During inference, these models resize the shortest side of images to the indicated scale, while preserving the aspect ratio. $^\ddagger$Our re-implementation. \textsuperscript{c}Models for these ADE20K results are pre-trained for COCO panoptic segmentation.
    }
    \label{tab:sota_panoptic}
\end{table*}

\begin{table*}[t]
    \centering
    \scriptsize
    \renewcommand{\tabcolsep}{3.2pt}
    \begin{tabularx}{\linewidth}
    {llcc c cYYY c cYYY}
        \toprule
        
        \multirow{2}[2]{*}{Method} &
        \multirow{2}[2]{*}{Backbone} &
        \multirow{2}[2]{*}{Pre-training} &
        \multirow{2}[2]{*}{Params} &&
        \multicolumn{4}{c}{Cityscapes \emph{val}~\cite{cordts2016cityscapes}} && 
        \multicolumn{4}{c}{ADE20K \emph{val}~\cite{zhou2017ade20k}} \\
        \cmidrule{6-9}
        \cmidrule{11-14}
        &
        &
        &
        &&
        Input size & GFLOPs & FPS & mIoU &&
        Input size & GFLOPs & FPS & mIoU \\

        \midrule

        Mask2Former$^\dagger$~\cite{cheng2022mask2former} & 
        Swin-L~\cite{liu2021swin} &
        IN21K &  
        216M &&
        $1024\times2048$ & -- & 14 & 83.3 &&
        $640^2$ & -- & 33 & 56.1  \\

        MaskDINO$^\dagger$~\cite{li2023maskdino} & 
        Swin-L~\cite{liu2021swin} & 
        IN21K & 
        223M &&
        -- & -- & -- & -- &&
        $640^2$ & -- & -- & 56.6   \\
        
        OneFormer$^\dagger$~\cite{jain2023oneformer} & 
        ConvNext-XL~\cite{liu2022convnext} & 
        IN21K & 
        373M &&
        $1024\times2048$ & 775 & 
        7
        & 83.6 &&
        $640^2$ & 607 & 21 & 57.4  \\
        
        OneFormer$^\dagger$~\cite{jain2023oneformer} & 
        DiNAT-L~\cite{hassani2022dinat} & 
        IN21K & 
        223M &&
        $1024\times2048$ & 450 & 14 & 83.1 &&
        $896^2$ & 678 & 19 & 58.1   \\

        kMaX-DeepLab~\cite{yu2022kmaxdeeplab}  & 
        ConvNext-L~\cite{liu2022convnext} & 
        IN21K & 
        232M &&
        $1025\times2049$  & 1673 & -- & 83.5  &&
        -- & -- & -- & --   \\
        
        Mask2Former~\cite{cheng2022mask2former} & 
        ViT-L~\cite{dosovitskiy2021vit} &
        DINOv2 + DA &  
        -- &&
        $896\times1792$ & -- & -- & 84.8 &&
        $896^2$ & -- & -- & 59.4  \\
        
        \midrule
        
        Mask2Former$^\ddagger$~\cite{cheng2022mask2former} & 
        ViT-Adapter-L$^\ddagger$~\cite{chen2023vitadapter} & 
        DINOv2 & 
        351M &&
        $1024^2$ & 5200 & 
        7
        & 84.5 &&
        $512^2$ & 910 & 21 & 58.9 \\
        
        \midrule
        
        \textbf{EoMT} (Ours) & 
        ViT-L~\cite{dosovitskiy2021vit} & 
        DINOv2 & 
        319M &&
        $1024^2$ & 4350 & 25 & 84.2 &&
        $512^2$ & 721 & 92 & 58.4 \\

        \bottomrule
    \end{tabularx}
    \caption{\textbf{EoMT for semantic segmentation.} $^\dagger$On ADE20K, these models resize the shortest side of images to the indicated scale during inference, while preserving the aspect ratio. $^\ddagger$Our re-implementation. ViT-Adapter + Mask2Former and EoMT use windowed inference, dividing each image into multiple crops, and the FLOPs and FPS results account for this. DA is Depth Anything~\cite{yang2024depthanything}.}
    \label{tab:sota_semantic}
\end{table*}

\PAR{Impact of model size.}
So far, we have only shown results for one model size, \ie, \mbox{ViT-L}. In \cref{tab:model_size}, we assess the impact of model size on the importance of task-specific components. The results show that the relative performance of EoMT compared to ViT-Adapter + M2F improves as the size of the model increases. While EoMT lags behind the more complex model by 5.8~PQ for \mbox{ViT-S}, the performance gap narrows significantly as the model scales up, becoming only 0.7~PQ for \mbox{ViT-g}. This shows that increasing the capacity of the ViT, combined with strong pre-training, allows EoMT to better solve the image segmentation task. This further confirms our hypothesis that the necessity for task-specific components decreases as the capacity of the model increases. At the same time, we also observe that replacing the complex ViT-Adapter + M2F model with EoMT significantly improves the prediction speed at all model sizes. Interestingly, this allows EoMT to use larger models and obtain higher scores than ViT-Adapter + M2F, while being significantly faster. For example, EoMT with \mbox{ViT-L} obtains a PQ of 56.0 at 128~FPS, which is both significantly faster and more accurate than ViT-Adapter + M2F with ViT-B, at a PQ of 54.4 at 32 FPS. As shown in \cref{fig:comparison}, EoMT obtains a better PQ~\vs~FPS trade-off across all model sizes that we tested. This shows the power of EoMT and its simplicity.

\PAR{EoMT on different benchmarks.}
To demonstrate EoMT’s versatility across image segmentation tasks and datasets, we evaluate its performance on multiple benchmarks. For panoptic segmentation, as shown in~\cref{tab:sota_panoptic}, EoMT achieves a significantly better PQ~\vs~FPS trade-off than ViT-Adapter + M2F~\cite{chen2023vitadapter, cheng2022mask2former} on COCO~\cite{lin2014coco}, and a similar trade-off on ADE20K~\cite{zhou2017ade20k}, while being significantly simpler. Moreover, on COCO, EoMT achieves a PQ that is on par with existing state-of-the-art methods while being up to $2.1\times$ faster, highlighting the strength of our simplified design. %

For semantic segmentation on Cityscapes~\cite{cordts2016cityscapes} and ADE20K, \cref{tab:sota_semantic} shows that EoMT performs comparably to the more complex ViT-Adapter + M2F baseline in terms of mIoU, while being considerably faster, \ie, up to $4.4\times$. Compared to other state-of-the-art methods, EoMT again obtains competitive performance, even though it is a much simpler and more efficient model. This shows that EoMT is also highly effective for semantic segmentation, providing a strong balance between speed and accuracy.

For instance segmentation on COCO, we see similar positive results in \cref{tab:sota_instance}. Although the overall accuracy drop is slightly higher than for the other tasks, EoMT still achieves a better AP~\vs~FPS trade-off than ViT-Adapter + M2F, \eg, 48.8~AP at 30~FPS \vs 47.6~AP at 29~FPS. Overall, these results demonstrate the strength and general applicability of EoMT, as it performs effectively across a variety of segmentation tasks and datasets.

\PAR{Importance of mask annealing.}
In~\cref{tab:steps}, we observed that mask annealing allows EoMT to remove masked attention during inference to improve efficiency, while keeping the PQ roughly the same. In~\cref{tab:mask_annealing}, we compare mask annealing to alternative strategies. 
The first alternative approach of training with masked attention and simply disabling it during inference causes a severe performance drop, showing that using a different strategy at training and inference time is ineffective. The second alternative approach of disabling masked attention during both training and inference does not fail catastrophically, but it results in a PQ that is still significantly lower than when using masked attention. In contrast, mask annealing enables EoMT to leverage masked attention during early training stages to improve convergence and roughly maintain the PQ, while eliminating the need for masking during inference, thereby more than doubling inference speed. As such, it is a key component of EoMT. More results on the general applicability of mask annealing are provided in
\cref{sec:supp:exp} and \cref{tab:supp:mask_annealing_detail}.

\begin{table}
    \centering
    \notsotiny
    \renewcommand{\tabcolsep}{2.3pt}
    \begin{tabularx}{\linewidth}
    {llc ccccYY}
        \toprule
        Method & Backbone & PT & Params & Input  & GFLOPs & FPS & AP \\
        \midrule
        OneFormer$^\dagger$~\cite{jain2023oneformer} & 
        DiNAT-L~\cite{hassani2022dinat} & 
        I & 
        223M & $800^2$ & 736 & 20 & 49.2 \\
        
        Mask2Former$^\dagger$~\cite{cheng2022mask2former} & 
        Swin-L~\cite{liu2021swin} & 
        I & 
        216M & $800^2$ & 868 & 24 & 50.1 \\
        
        MaskDINO$^\dagger$~\cite{li2023maskdino} & 
        Swin-L~\cite{liu2021swin} & 
        I & 
        223M & $800^2$ & 1326 & 14 & 52.3 \\
        
        \midrule

        Mask2Former$^\ddagger$~\cite{cheng2022mask2former} & 
        ViT-Adapter-L$^\ddagger$~\cite{chen2023vitadapter} & 
        D & 
        349M & $640^2$ & 830 & 29 & 47.6 \\

        Mask2Former$^\ddagger$~\cite{cheng2022mask2former} & 
        ViT-Adapter-L$^\ddagger$~\cite{chen2023vitadapter} & 
        D & 
        354M & $1280^2$ & 4817 & 10 & 51.4\\
        
        \midrule

        \textbf{EoMT} (Ours) & 
        ViT-L~\cite{dosovitskiy2021vit} & 
        D & 
        316M & $640^2$& 669 & 128 & 45.2 \\
        
        \textbf{EoMT} (Ours) & 
        ViT-L~\cite{dosovitskiy2021vit} & 
        D & 
        322M & $1280^2$ & 4146 & 30 & 48.8 \\
        \bottomrule
    \end{tabularx}
    \caption{\textbf{EoMT for instance segmentation.} Evaluated on COCO \textit{val2017}~\cite{lin2014coco}. $^\dagger$During inference, these models resize the shortest side of images to the indicated scale while preserving the aspect ratio. $^\ddagger$Our re-implementation. \textbf{PT} indicates pre-training (\textbf{I} for ImageNet-21K~\cite{deng2009imagenet} and \textbf{D} for DINOv2~\cite{oquab2023dinov2}).
    }
    \label{tab:sota_instance}
\end{table}

\begin{table}[t]
\setlength{\tabcolsep}{1.5pt}
\footnotesize
\centering
\begin{tabularx}{\linewidth}{clclcYY@{\hskip -4pt}c}
\toprule
\multicolumn{2}{l}{Training} & \multicolumn{2}{l}{Inference} &~~~& GFLOPs & FPS & PQ \\
\midrule

\rcmark & Masking &  %
\rcmark & Masking &&  %
828 & 
61 & 
\phantom{\qdecnew{0.00}}56.2\phantom{\qdecnew{0.00}} \\

\rcmark & Masking &  %
\gxmark & w/o Masking &&  %
669 & 
128 & 
\phantom{\qdecnew{0.00}}27.4\qdecnew{28.8} \\

\gxmark & w/o Masking &  %
\gxmark & w/o Masking &&  %
669 & 
128 & 
\phantom{\qdecnew{0.00}}53.2\qdecnew{3.0\phantom{0}} \\

\rcmark \fancyrightarrow \gxmark&\textit{Mask annealing} &  %
\gxmark & w/o Masking &&  %
669 & 
128 & 
\phantom{\qdecnew{0.00}}56.0\qdecnew{0.2\phantom{0}} \\

\bottomrule
\end{tabularx}
\caption{\textbf{\textit{Mask annealing.}} Effectively removes masked attention during inference. When never masking, intermediate masks are not predicted or supervised. Evaluated on COCO \textit{val2017}~\cite{lin2014coco}. %
}
\label{tab:mask_annealing}
\end{table}

\subsection{Further Experiments}
\label{sec:exp:further}

\PARbegin{Out-of-distribution generalization.} Since EoMT is ViT-based, it supports initialization with vision foundation model (VFMs) like DINOv2~\cite{oquab2023dinov2}, which achieve state-of-the-art out-of-distribution (OOD) generalization~\cite{englert2024vfmuda, kerssies2024evaluating}. In contrast, prior segmentation models typically rely on ConvNeXt~\cite{liu2022convnext}, Swin~\cite{liu2021swin}, or other non-plain ViT backbones, which cannot leverage VFM pre-training due to their incompatible architectures. We evaluate OOD generalization on the BRAVO~\cite{vu2024bravo} benchmark, by training on Cityscapes and evaluating on multiple OOD datasets. \cref{tab:ood} shows that DINOv2-based models demonstrate superior OOD generalization, outperforming the Swin-based model by more than 7.8~mIoU despite similar in-distribution performance on Cityscapes. Importantly, EoMT leverages the strong OOD performance of VFMs while being significantly more efficient than existing VFM-based methods, which is essential for real-world applications. Further analysis of OOD confidence estimation, showing that EoMT produces significantly more reliable confidence scores than ViT-Adapter + M2F, is provided in 
\cref{sec:supp:ood}.

\PAR{Token merging.}
EoMT benefits from ongoing ViT advancements by using the plain ViT architecture. One such advancement is token merging~\cite{bolya2023tome,norouzi2024algm,lu2023cts}, which improves efficiency by merging semantically redundant tokens while preserving segmentation accuracy. As shown in \cref{tab:algm}, EoMT is compatible with ALGM~\cite{norouzi2024algm}, the state-of-the-art token merging method for semantic segmentation. With ALGM, EoMT’s throughput increases by up to $31\%$ without affecting mIoU. In contrast, while ViT-Adapter + M2F reduces in FLOPs, it sees no throughput gain, as it is bottlenecked by its additional components and the overhead of ‘unmerging’ tokens for ViT-Adapter interaction. This highlights the benefit of keeping EoMT close to the plain ViT.

\begin{table}
    \centering
    \footnotesize
    \setlength{\tabcolsep}{2.0pt}
    \begin{tabularx}{\linewidth}{
    llcYY
    }
    \toprule

    Method & Backbone & Pre-training & mIoU\textsubscript{ID} & mIoU\textsubscript{OOD} \\

    \midrule

    M2F~\cite{cheng2022mask2former} &
    Swin-L~\cite{liu2021swin} &
    IN21K &
    83.3 &
    69.4 \\

    M2F$^\ddagger$~\cite{cheng2022mask2former} &
    ViT-Adapter-L$^\ddagger$~\cite{chen2023vitadapter} &
    DINOv2 &
    84.5 &
    78.0 \\

    \textbf{EoMT} (Ours) &
    ViT-L~\cite{dosovitskiy2021vit} &
    DINOv2 &
    84.2 &
    77.2 \\

    \bottomrule
    
    \end{tabularx}
    \caption{\textbf{Out-of-distribution generalization.} Despite similar in-distribution performance on Cityscapes \textit{val} (mIoU\textsubscript{ID}), DINOv2-based models generalize significantly better out-of-distribution (mIoU\textsubscript{OOD}). Trained on Cityscapes \textit{train}~\cite{cordts2016cityscapes}, evaluated on BRAVO~\cite{vu2024bravo}. $^\ddagger$Our re-implementation. 
    }
    \label{tab:ood}
\end{table}

\begin{table}[t]
    \centering
    \footnotesize
    \setlength{\tabcolsep}{2.5pt}
    \begin{tabularx}{\linewidth}{
    lcYcY
    }
    \toprule
    
    Method & 
    Token merging & 
    GFLOPs & 
    Throughput & 
    mIoU \\

    \midrule

    \multirow{2}{*}{\shortstack[l]{ViT-Adapter + M2F}} &
    \xmark &
    5200 &
    \phantom{\qincnew{2\%}}9\phantom{\qincnew{2\%}} & 
    84.5 \\

    &
    \cmark &
    3031 &
    \phantom{\qincnew{2\%}}9\qincnew{0\%} & 
    84.3 \\

    \midrule

    \multirow{2}{*}{\textbf{EoMT} (Ours)} &
    \xmark &
    4350 &
    \phantom{\qincnew{31\%}}29\phantom{\qincnew{31\%}} & 
    84.2 \\

    &
    \cmark &
    1183 &
    \phantom{\qincnew{31\%}}38\qincnew{31\%} & %
    84.2 \\

    \bottomrule

    \end{tabularx}
    \caption{\textbf{Token merging.} EoMT’s ViT-only design is no longer bottlenecked by additional task-specific components, enabling ViT optimizations like ALGM~\cite{norouzi2024algm}. Throughput is in \textit{images per second}, with a batch size of 32. Evaluated on Cityscapes \textit{val}~\cite{cordts2016cityscapes}.
    }
    \label{tab:algm}
\end{table}

\section{Conclusion}
\label{sec:conclusion}
In this paper, we show that task-specific components for image segmentation with Vision Transformers (ViTs) become increasingly redundant as model size and pre-training are scaled up. By removing all such components, we introduce the Encoder-only Mask Transformer (EoMT), a segmentation model that purely uses a plain ViT, revealing that \textit{your ViT is secretly an image segmentation model}. %
EoMT delivers both high accuracy and impressive speed, with a significantly simpler design than existing models. Our findings indicate that compute resources should not be spent on adding architectural complexity, but rather on scaling the ViT and pre-training, as we found that these factors are critical in improving performance.
As a simple and scalable approach, EoMT provides a solid foundation for next-generation segmentation models that readily adapts to advances in the rapidly evolving fields of Transformers and foundation models.

\footnotesize 
\PAR{Acknowledgements.}
This work was supported by Chips Joint Undertaking (Chips JU) in EdgeAI ``Edge AI Technologies for Optimised Performance Embedded Processing'' project, grant agreement no.~101097300.
Niccolò Cavagnero acknowledges travel support from the European Union’s Horizon 2020 research and innovation program, grant agreement no.~951847.
Giuseppe Averta was supported by FAIR - Future Artificial Intelligence Research, Next-GenEU (PNRR – MISS. 4 COMP. 2, INV. 1.3 – D.D. 1555 11/10/2022, PE00000013).
We also acknowledge the CINECA award under the ISCRA initiative and the Dutch national e-infrastructure with the support of the SURF Cooperative, grant agreement no.~EINF-9663 and EINF-11151, financed by the Dutch Research Council (NWO), for the availability of high-performance computing resources and support.

{
    \small
    \bibliographystyle{ieeenat_fullname}
    \bibliography{main}

\begin{thebibliography}{71}
\providecommand{\natexlab}[1]{#1}
\providecommand{\url}[1]{\texttt{#1}}
\expandafter\ifx\csname urlstyle\endcsname\relax
  \providecommand{\doi}[1]{doi: #1}\else
  \providecommand{\doi}{doi: \begingroup \urlstyle{rm}\Url}\fi

\bibitem[Alabdulmohsin et~al.(2023)Alabdulmohsin, Zhai, Kolesnikov, and Beyer]{alabdulmohsin2023sovit}
Ibrahim Alabdulmohsin, Xiaohua Zhai, Alexander Kolesnikov, and Lucas Beyer.
\newblock {Getting ViT in Shape: Scaling Laws for Compute-Optimal Model Design}.
\newblock In \emph{NeurIPS}, 2023.

\bibitem[Ansel et~al.(2024)Ansel, Yang, He, Gimelshein, Jain, Voznesensky, Bao, Bell, Berard, Burovski, Chauhan, Chourdia, Constable, Desmaison, DeVito, Ellison, Feng, Gong, Gschwind, Hirsh, Huang, Kalambarkar, Kirsch, Lazos, Lezcano, Liang, Liang, Lu, Luk, Maher, et~al.]{ansel2024pytorch2}
Jason Ansel, Edward Yang, Horace He, Natalia Gimelshein, Animesh Jain, Michael Voznesensky, Bin Bao, Peter Bell, David Berard, Evgeni Burovski, Geeta Chauhan, Anjali Chourdia, Will Constable, Alban Desmaison, Zachary DeVito, Elias Ellison, Will Feng, Jiong Gong, Michael Gschwind, Brian Hirsh, Sherlock Huang, Kshiteej Kalambarkar, Laurent Kirsch, Michael Lazos, Mario Lezcano, Yanbo Liang, Jason Liang, Yinghai Lu, C.~K. Luk, Bert Maher, et~al.
\newblock {PyTorch 2: Faster Machine Learning Through Dynamic Python Bytecode Transformation and Graph Compilation}.
\newblock In \emph{ASPLOS}, 2024.

\bibitem[Athar et~al.(2023)Athar, Hermans, Luiten, Ramanan, and Leibe]{athar2023tarvis}
Ali Athar, Alexander Hermans, Jonathon Luiten, Deva Ramanan, and Bastian Leibe.
\newblock {TarViS: A Unified Approach for Target-Based Video Segmentation}.
\newblock In \emph{CVPR}, 2023.

\bibitem[Ba et~al.(2016)Ba, Kiros, and Hinton]{ba2016layernorm}
Jimmy~Lei Ba, Jamie~Ryan Kiros, and Geoffrey~E. Hinton.
\newblock {Layer Normalization}.
\newblock \emph{arXiv preprint arXiv:1607.06450}, 2016.

\bibitem[Bao et~al.(2022)Bao, Dong, Piao, and Wei]{bao2022beit}
Hangbo Bao, Li Dong, Songhao Piao, and Furu Wei.
\newblock {BEiT: BERT Pre-Training of Image Transformers}.
\newblock In \emph{ICLR}, 2022.

\bibitem[Beyer et~al.(2023)Beyer, Izmailov, Kolesnikov, Caron, Kornblith, Zhai, Minderer, Tschannen, Alabdulmohsin, and Pavetic]{beyer2023flexivit}
Lucas Beyer, Pavel Izmailov, Alexander Kolesnikov, Mathilde Caron, Simon Kornblith, Xiaohua Zhai, Matthias Minderer, Michael Tschannen, Ibrahim Alabdulmohsin, and Filip Pavetic.
\newblock {FlexiViT: One Model for All Patch Sizes}.
\newblock In \emph{CVPR}, 2023.

\bibitem[Bolya et~al.(2023)Bolya, Fu, Dai, Zhang, Feichtenhofer, and Hoffman]{bolya2023tome}
Daniel Bolya, Cheng-Yang Fu, Xiaoliang Dai, Peizhao Zhang, Christoph Feichtenhofer, and Judy Hoffman.
\newblock {Token Merging: Your ViT But Faster}.
\newblock In \emph{ICLR}, 2023.

\bibitem[Carion et~al.(2020)Carion, Massa, Synnaeve, Usunier, Kirillov, and Zagoruyko]{carion2020end}
Nicolas Carion, Francisco Massa, Gabriel Synnaeve, Nicolas Usunier, Alexander Kirillov, and Sergey Zagoruyko.
\newblock {End-to-End Object Detection with Transformers}.
\newblock In \emph{ECCV}, 2020.

\bibitem[Caron et~al.(2021)Caron, Touvron, Misra, J{\'e}gou, Mairal, Bojanowski, and Joulin]{caron2021dino}
Mathilde Caron, Hugo Touvron, Ishan Misra, Herv{\'e} J{\'e}gou, Julien Mairal, Piotr Bojanowski, and Armand Joulin.
\newblock {Emerging Properties in Self-Supervised Vision Transformers}.
\newblock In \emph{ICCV}, 2021.

\bibitem[Cavagnero et~al.(2024)Cavagnero, Rosi, Cuttano, Pistilli, Ciccone, Averta, and Cermelli]{cavagnero2024pem}
Niccol{\`o} Cavagnero, Gabriele Rosi, Claudia Cuttano, Francesca Pistilli, Marco Ciccone, Giuseppe Averta, and Fabio Cermelli.
\newblock {PEM: Prototype-based Efficient MaskFormer for Image Segmentation}.
\newblock In \emph{CVPR}, 2024.

\bibitem[Chen et~al.(2018)Chen, Zhu, Papandreou, Schroff, and Adam]{chen2018deeplabv3plus}
Liang-Chieh Chen, Yukun Zhu, George Papandreou, Florian Schroff, and Hartwig Adam.
\newblock {Encoder-Decoder with Atrous Separable Convolution for Semantic Image Segmentation}.
\newblock In \emph{ECCV}, 2018.

\bibitem[Chen et~al.(2022)Chen, Du, Yang, Beyer, Zhai, Lin, Chen, Li, Song, Wang, et~al.]{chen2022simple}
Wuyang Chen, Xianzhi Du, Fan Yang, Lucas Beyer, Xiaohua Zhai, Tsung-Yi Lin, Huizhong Chen, Jing Li, Xiaodan Song, Zhangyang Wang, et~al.
\newblock {A Simple Single-Scale Vision Transformer for Object Localization and Instance Segmentation}.
\newblock In \emph{ECCV}, 2022.

\bibitem[Chen et~al.(2023)Chen, Duan, Wang, He, Lu, Dai, and Qiao]{chen2023vitadapter}
Zhe Chen, Yuchen Duan, Wenhai Wang, Junjun He, Tong Lu, Jifeng Dai, and Yu Qiao.
\newblock {Vision Transformer Adapter for Dense Predictions}.
\newblock In \emph{ICLR}, 2023.

\bibitem[Cheng et~al.(2021)Cheng, Schwing, and Kirillov]{cheng2021maskformer}
Bowen Cheng, Alex Schwing, and Alexander Kirillov.
\newblock {Per-Pixel Classification is Not All You Need for Semantic Segmentation}.
\newblock In \emph{NeurIPS}, 2021.

\bibitem[Cheng et~al.(2022)Cheng, Misra, Schwing, Kirillov, and Girdhar]{cheng2022mask2former}
Bowen Cheng, Ishan Misra, Alexander~G Schwing, Alexander Kirillov, and Rohit Girdhar.
\newblock {Masked-attention Mask Transformer for Universal Image Segmentation}.
\newblock In \emph{CVPR}, 2022.

\bibitem[Choquette(2023)]{choquette2023h100}
Jack Choquette.
\newblock {NVIDIA Hopper H100 GPU: Scaling Performance}.
\newblock \emph{IEEE Micro}, 43\penalty0 (3):\penalty0 9--17, 2023.

\bibitem[Cordts et~al.(2016)Cordts, Omran, Ramos, Rehfeld, Enzweiler, Benenson, Franke, Roth, and Schiele]{cordts2016cityscapes}
Marius Cordts, Mohamed Omran, Sebastian Ramos, Timo Rehfeld, Markus Enzweiler, Rodrigo Benenson, Uwe Franke, Stefan Roth, and Bernt Schiele.
\newblock {The Cityscapes Dataset for Semantic Urban Scene Understanding}.
\newblock In \emph{CVPR}, 2016.

\bibitem[Dao(2024)]{dao2024flashattention2}
Tri Dao.
\newblock {FlashAttention-2: Faster Attention with Better Parallelism and Work Partitioning}.
\newblock In \emph{ICLR}, 2024.

\bibitem[Dao et~al.(2022)Dao, Fu, Ermon, Rudra, and Re]{dao2022flashattention}
Tri Dao, Daniel~Y Fu, Stefano Ermon, Atri Rudra, and Christopher Re.
\newblock {FlashAttention: Fast and Memory-Efficient Exact Attention with {IO}-Awareness}.
\newblock In \emph{NeurIPS}, 2022.

\bibitem[Darcet et~al.(2024)Darcet, Oquab, Mairal, and Bojanowski]{darcet2024registers}
Timoth{\'e}e Darcet, Maxime Oquab, Julien Mairal, and Piotr Bojanowski.
\newblock {Vision Transformers Need Registers}.
\newblock In \emph{ICLR}, 2024.

\bibitem[Dehghani et~al.(2023)Dehghani, Djolonga, Mustafa, Padlewski, Heek, Gilmer, Steiner, Caron, Geirhos, Alabdulmohsin, Jenatton, Beyer, Tschannen, Arnab, Wang, Riquelme~Ruiz, Minderer, Puigcerver, Evci, Kumar, Steenkiste, Elsayed, Mahendran, Yu, Oliver, Huot, Bastings, Collier, Gritsenko, Birodkar, Vasconcelos, Tay, Mensink, Kolesnikov, Pavetic, Tran, Kipf, Lucic, Zhai, Keysers, Harmsen, and Houlsby]{dehghani2023vit22b}
Mostafa Dehghani, Josip Djolonga, Basil Mustafa, Piotr Padlewski, Jonathan Heek, Justin Gilmer, Andreas~Peter Steiner, Mathilde Caron, Robert Geirhos, Ibrahim Alabdulmohsin, Rodolphe Jenatton, Lucas Beyer, Michael Tschannen, Anurag Arnab, Xiao Wang, Carlos Riquelme~Ruiz, Matthias Minderer, Joan Puigcerver, Utku Evci, Manoj Kumar, Sjoerd~Van Steenkiste, Gamaleldin~Fathy Elsayed, Aravindh Mahendran, Fisher Yu, Avital Oliver, Fantine Huot, Jasmijn Bastings, Mark Collier, Alexey~A. Gritsenko, Vighnesh Birodkar, Cristina~Nader Vasconcelos, Yi Tay, Thomas Mensink, Alexander Kolesnikov, Filip Pavetic, Dustin Tran, Thomas Kipf, Mario Lucic, Xiaohua Zhai, Daniel Keysers, Jeremiah~J. Harmsen, and Neil Houlsby.
\newblock {Scaling Vision Transformers to 22 Billion Parameters}.
\newblock In \emph{ICML}, 2023.

\bibitem[Deng et~al.(2009)Deng, Dong, Socher, Li, Li, and Fei-Fei]{deng2009imagenet}
Jia Deng, Wei Dong, Richard Socher, Li-Jia Li, Kai Li, and Li Fei-Fei.
\newblock {ImageNet: A large-scale hierarchical image database}.
\newblock In \emph{CVPR}, 2009.

\bibitem[Dosovitskiy et~al.(2021)Dosovitskiy, Beyer, Kolesnikov, Weissenborn, Zhai, Unterthiner, Dehghani, Minderer, Heigold, Gelly, Uszkoreit, and Houlsby]{dosovitskiy2021vit}
Alexey Dosovitskiy, Lucas Beyer, Alexander Kolesnikov, Dirk Weissenborn, Xiaohua Zhai, Thomas Unterthiner, Mostafa Dehghani, Matthias Minderer, Georg Heigold, Sylvain Gelly, Jakob Uszkoreit, and Neil Houlsby.
\newblock {An Image is Worth 16x16 Words: Transformers for Image Recognition at Scale}.
\newblock In \emph{ICLR}, 2021.

\bibitem[Elster and Haugdahl(2022)]{elster2022hopper}
Anne~C. Elster and Tor~A. Haugdahl.
\newblock {Nvidia Hopper GPU and Grace CPU Highlights}.
\newblock \emph{Computing in Science \& Engineering}, 24\penalty0 (2):\penalty0 95--100, 2022.

\bibitem[Englert et~al.(2024)Englert, Piva, Kerssies, De~Geus, and Dubbelman]{englert2024vfmuda}
Brun\'o~B. Englert, Fabrizio~J. Piva, Tommie Kerssies, Daan De~Geus, and Gijs Dubbelman.
\newblock {Exploring the Benefits of Vision Foundation Models for Unsupervised Domain Adaptation}.
\newblock In \emph{CVPR Workshops}, 2024.

\bibitem[Everingham et~al.(2010)Everingham, Van~Gool, Williams, Winn, and Zisserman]{everingham2010pascal}
Mark Everingham, Luc Van~Gool, Christopher K.~I. Williams, John Winn, and Andrew Zisserman.
\newblock {The Pascal Visual Object Classes (VOC) Challenge}.
\newblock \emph{IJCV}, 2010.

\bibitem[Fang et~al.(2021)Fang, Liao, Wang, Fang, Qi, Wu, Niu, and Liu]{fang2021you}
Yuxin Fang, Bencheng Liao, Xinggang Wang, Jiemin Fang, Jiyang Qi, Rui Wu, Jianwei Niu, and Wenyu Liu.
\newblock {You Only Look at One Sequence: Rethinking Transformer in Vision through Object Detection}.
\newblock In \emph{NeurIPS}, 2021.

\bibitem[Fang et~al.(2024)Fang, Sun, Wang, Huang, Wang, and Cao]{fang2024eva02}
Yuxin Fang, Quan Sun, Xinggang Wang, Tiejun Huang, Xinlong Wang, and Yue Cao.
\newblock {EVA-02: A Visual Representation for Neon Genesis}.
\newblock \emph{Image and Vision Computing}, 2024.

\bibitem[Ghiasi et~al.(2021)Ghiasi, Cui, Srinivas, Qian, Lin, Cubuk, Le, and Zoph]{ghiasi2021copypaste}
Golnaz Ghiasi, Yin Cui, Aravind Srinivas, Rui Qian, Tsung-Yi Lin, Ekin~D Cubuk, Quoc~V Le, and Barret Zoph.
\newblock {Simple Copy-Paste is a Strong Data Augmentation Method for Instance Segmentation}.
\newblock In \emph{CVPR}, 2021.

\bibitem[Hassani and Shi(2022)]{hassani2022dinat}
Ali Hassani and Humphrey Shi.
\newblock {Dilated Neighborhood Attention Transformer}.
\newblock \emph{arXiv preprint arXiv:2209.15001}, 2022.

\bibitem[Hassani et~al.(2023)Hassani, Walton, Li, Li, and Shi]{hassani2023nat}
Ali Hassani, Steven Walton, Jiachen Li, Shen Li, and Humphrey Shi.
\newblock {Neighborhood Attention Transformer}.
\newblock In \emph{CVPR}, 2023.

\bibitem[He et~al.(2016)He, Zhang, Ren, and Sun]{he2016resnet}
Kaiming He, Xiangyu Zhang, Shaoqing Ren, and Jian Sun.
\newblock {Deep Residual Learning for Image Recognition}.
\newblock In \emph{CVPR}, 2016.

\bibitem[He et~al.(2017)He, Gkioxari, Dollar, and Girshick]{he2017maskrcnn}
Kaiming He, Georgia Gkioxari, Piotr Dollar, and Ross Girshick.
\newblock {Mask R-CNN}.
\newblock In \emph{ICCV}, 2017.

\bibitem[Hu et~al.(2023)Hu, Huang, Ren, Zhang, Ji, and Cao]{hu2023yoso}
Jie Hu, Linyan Huang, Tianhe Ren, Shengchuan Zhang, Rongrong Ji, and Liujuan Cao.
\newblock {You Only Segment Once: Towards Real-Time Panoptic Segmentation}.
\newblock In \emph{CVPR}, 2023.

\bibitem[Jain et~al.(2023)Jain, Li, Chiu, Hassani, Orlov, and Shi]{jain2023oneformer}
Jitesh Jain, Jiachen Li, Mang~Tik Chiu, Ali Hassani, Nikita Orlov, and Humphrey Shi.
\newblock {OneFormer: One Transformer To Rule Universal Image Segmentation}.
\newblock In \emph{CVPR}, 2023.

\bibitem[Kerssies et~al.(2024{\natexlab{a}})Kerssies, De~Geus, and Dubbelman]{kerssies2024benchmark}
Tommie Kerssies, Daan De~Geus, and Gijs Dubbelman.
\newblock {How to Benchmark Vision Foundation Models for Semantic Segmentation?}
\newblock In \emph{CVPR Workshops}, 2024{\natexlab{a}}.

\bibitem[Kerssies et~al.(2024{\natexlab{b}})Kerssies, de~Geus, and Dubbelman]{kerssies2024evaluating}
Tommie Kerssies, Daan de Geus, and Gijs Dubbelman.
\newblock {First Place Solution to the ECCV 2024 BRAVO Challenge: Evaluating Robustness of Vision Foundation Models for Semantic Segmentation}.
\newblock \emph{arXiv preprint arXiv:2409.17208}, 2024{\natexlab{b}}.

\bibitem[Kirillov et~al.(2019{\natexlab{a}})Kirillov, Girshick, He, and Dollar]{kirillov2019panopticfpn}
Alexander Kirillov, Ross Girshick, Kaiming He, and Piotr Dollar.
\newblock {Panoptic Feature Pyramid Networks}.
\newblock In \emph{CVPR}, 2019{\natexlab{a}}.

\bibitem[Kirillov et~al.(2019{\natexlab{b}})Kirillov, He, Girshick, Rother, and Doll{\'a}r]{kirillov2019panoptic}
Alexander Kirillov, Kaiming He, Ross Girshick, Carsten Rother, and Piotr Doll{\'a}r.
\newblock {Panoptic Segmentation}.
\newblock In \emph{CVPR}, 2019{\natexlab{b}}.

\bibitem[Li et~al.(2023)Li, Zhang, Xu, Liu, Zhang, Ni, and Shum]{li2023maskdino}
Feng Li, Hao Zhang, Huaizhe Xu, Shilong Liu, Lei Zhang, Lionel~M. Ni, and Heung-Yeung Shum.
\newblock {Mask DINO: Towards a Unified Transformer-Based Framework for Object Detection and Segmentation}.
\newblock In \emph{CVPR}, 2023.

\bibitem[Li et~al.(2024)Li, Yuan, Li, Ding, Wu, Zhang, Li, Chen, and Loy]{OMGSeg}
Xiangtai Li, Haobo Yuan, Wei Li, Henghui Ding, Size Wu, Wenwei Zhang, Yining Li, Kai Chen, and Chen~Change Loy.
\newblock Omg-seg: Is one model good enough for all segmentation?
\newblock In \emph{CVPR}, 2024.

\bibitem[Li et~al.(2022)Li, Mao, Girshick, and He]{li2022vitdet}
Yanghao Li, Hanzi Mao, Ross Girshick, and Kaiming He.
\newblock {Exploring Plain Vision Transformer Backbones for Object Detection}.
\newblock In \emph{ECCV}, 2022.

\bibitem[Lin et~al.(2014)Lin, Maire, Belongie, Hays, Perona, Ramanan, Doll{\'a}r, and Zitnick]{lin2014coco}
Tsung-Yi Lin, Michael Maire, Serge Belongie, James Hays, Pietro Perona, Deva Ramanan, Piotr Doll{\'a}r, and C~Lawrence Zitnick.
\newblock {Microsoft COCO: Common Objects in Context}.
\newblock In \emph{ECCV}, 2014.

\bibitem[Lin et~al.(2017)Lin, Dollar, Girshick, He, Hariharan, and Belongie]{lin2017fpn}
Tsung-Yi Lin, Piotr Dollar, Ross Girshick, Kaiming He, Bharath Hariharan, and Serge Belongie.
\newblock {Feature Pyramid Networks for Object Detection}.
\newblock In \emph{CVPR}, 2017.

\bibitem[Liu et~al.(2021)Liu, Lin, Cao, Hu, Wei, Zhang, Lin, and Guo]{liu2021swin}
Ze Liu, Yutong Lin, Yue Cao, Han Hu, Yixuan Wei, Zheng Zhang, Stephen Lin, and Baining Guo.
\newblock {Swin Transformer: Hierarchical Vision Transformer Using Shifted Windows}.
\newblock In \emph{ICCV}, 2021.

\bibitem[Liu et~al.(2022)Liu, Mao, Wu, Feichtenhofer, Darrell, and Xie]{liu2022convnext}
Zhuang Liu, Hanzi Mao, Chao-Yuan Wu, Christoph Feichtenhofer, Trevor Darrell, and Saining Xie.
\newblock {A ConvNet for the 2020s}.
\newblock In \emph{CVPR}, 2022.

\bibitem[Loshchilov and Hutter(2019)]{loshchilov2019adamw}
Ilya Loshchilov and Frank Hutter.
\newblock {Decoupled Weight Decay Regularization}.
\newblock In \emph{ICLR}, 2019.

\bibitem[Lu et~al.(2023)Lu, de~Geus, and Dubbelman]{lu2023cts}
Chenyang Lu, Daan de Geus, and Gijs Dubbelman.
\newblock {Content-Aware Token Sharing for Efficient Semantic Segmentation With Vision Transformers}.
\newblock In \emph{CVPR}, 2023.

\bibitem[Milletari et~al.(2016)Milletari, Navab, and Ahmadi]{milletari2016vnet}
Fausto Milletari, Nassir Navab, and Seyed-Ahmad Ahmadi.
\newblock {V-Net: Fully Convolutional Neural Networks for Volumetric Medical Image Segmentation}.
\newblock In \emph{3DV}, 2016.

\bibitem[Norouzi et~al.(2024)Norouzi, Orlova, de~Geus, and Dubbelman]{norouzi2024algm}
Narges Norouzi, Svetlana Orlova, Daan de Geus, and Gijs Dubbelman.
\newblock {ALGM: Adaptive Local-then-Global Token Merging for Efficient Semantic Segmentation with Plain Vision Transformers}.
\newblock In \emph{CVPR}, 2024.

\bibitem[Oquab et~al.(2024)Oquab, Darcet, Moutakanni, Vo, Szafraniec, Khalidov, Fernandez, Haziza, Massa, El-Nouby, et~al.]{oquab2023dinov2}
Maxime Oquab, Timoth{\'e}e Darcet, Th{\'e}o Moutakanni, Huy Vo, Marc Szafraniec, Vasil Khalidov, Pierre Fernandez, Daniel Haziza, Francisco Massa, Alaaeldin El-Nouby, et~al.
\newblock {DINOv2: Learning Robust Visual Features without Supervision}.
\newblock \emph{TMLR}, 2024.

\bibitem[Radford et~al.(2021)Radford, Kim, Hallacy, Ramesh, Goh, Agarwal, Sastry, Askell, Mishkin, Clark, et~al.]{radford2021clip}
Alec Radford, Jong~Wook Kim, Chris Hallacy, Aditya Ramesh, Gabriel Goh, Sandhini Agarwal, Girish Sastry, Amanda Askell, Pamela Mishkin, Jack Clark, et~al.
\newblock {Learning Transferable Visual Models From Natural Language Supervision}.
\newblock In \emph{ICML}, 2021.

\bibitem[Research(2023)]{meta2023fvcore}
Meta Research.
\newblock {fvcore}, 2023.

\bibitem[Russakovsky et~al.(2015)Russakovsky, Deng, Su, Krause, Satheesh, Ma, Huang, Karpathy, Khosla, Bernstein, et~al.]{russakovsky2015imagenet}
Olga Russakovsky, Jia Deng, Hao Su, Jonathan Krause, Sanjeev Satheesh, Sean Ma, Zhiheng Huang, Andrej Karpathy, Aditya Khosla, Michael Bernstein, et~al.
\newblock {ImageNet Large Scale Visual Recognition Challenge}.
\newblock \emph{IJCV}, 2015.

\bibitem[Strudel et~al.(2021)Strudel, Garcia, Laptev, and Schmid]{strudel2021segmenter}
Robin Strudel, Ricardo Garcia, Ivan Laptev, and Cordelia Schmid.
\newblock {Segmenter: Transformer for Semantic Segmentation}.
\newblock In \emph{ICCV}, 2021.

\bibitem[Touvron et~al.(2022)Touvron, Cord, and J{\'e}gou]{touvron2022deit}
Hugo Touvron, Matthieu Cord, and Herv{\'e} J{\'e}gou.
\newblock {DeiT III: Revenge of the ViT}.
\newblock In \emph{ECCV}, 2022.

\bibitem[Vaswani et~al.(2017)Vaswani, Shazeer, Parmar, Uszkoreit, Jones, Gomez, Kaiser, and Polosukhin]{vaswani2017transformer}
Ashish Vaswani, Noam Shazeer, Niki Parmar, Jakob Uszkoreit, Llion Jones, Aidan~N Gomez, \L~ukasz Kaiser, and Illia Polosukhin.
\newblock {Attention is All you Need}.
\newblock In \emph{NeurIPS}, 2017.

\bibitem[Vu et~al.(2024)Vu, Valle, Bursuc, Kerssies, de~Geus, Dubbelman, Qian, Zhu, Chen, Tang, Wang, Vojíř, Šochman, Matas, Smith, Ferrie, Basu, Sakaridis, and Van~Gool]{vu2024bravo}
Tuan-Hung Vu, Eduardo Valle, Andrei Bursuc, Tommie Kerssies, Daan de Geus, Gijs Dubbelman, Long Qian, Bingke Zhu, Yingying Chen, Ming Tang, Jinqiao Wang, Tomáš Vojíř, Jan Šochman, Jiří Matas, Michael Smith, Frank Ferrie, Shamik Basu, Christos Sakaridis, and Luc Van~Gool.
\newblock {The BRAVO Semantic Segmentation Challenge Results in UNCV2024}.
\newblock In \emph{ECCV Workshops}, 2024.

\bibitem[Wang et~al.(2021)Wang, Zhu, Adam, Yuille, and Chen]{wang2021max}
Huiyu Wang, Yukun Zhu, Hartwig Adam, Alan Yuille, and Liang-Chieh Chen.
\newblock {MaX-DeepLab: End-to-End Panoptic Segmentation with Mask Transformers}.
\newblock In \emph{CVPR}, 2021.

\bibitem[Wightman(2019)]{rw2019timm}
Ross Wightman.
\newblock {PyTorch Image Models}, 2019.

\bibitem[Wolf et~al.(2020)Wolf, Debut, Sanh, Chaumond, Delangue, Moi, Cistac, Rault, Louf, Funtowicz, Davison, Shleifer, von Platen, Ma, Jernite, Plu, Xu, Scao, Gugger, Drame, Lhoest, and Rush]{wolf-etal-2020-transformers}
Thomas Wolf, Lysandre Debut, Victor Sanh, Julien Chaumond, Clement Delangue, Anthony Moi, Pierric Cistac, Tim Rault, Rémi Louf, Morgan Funtowicz, Joe Davison, Sam Shleifer, Patrick von Platen, Clara Ma, Yacine Jernite, Julien Plu, Canwen Xu, Teven~Le Scao, Sylvain Gugger, Mariama Drame, Quentin Lhoest, and Alexander~M. Rush.
\newblock {Transformers: State-of-the-Art Natural Language Processing}.
\newblock In \emph{EMNLP Demos}, 2020.

\bibitem[Wu et~al.(2019)Wu, Kirillov, Massa, Lo, and Girshick]{wu2019detectron2}
Yuxin Wu, Alexander Kirillov, Francisco Massa, Wan-Yen Lo, and Ross Girshick.
\newblock {Detectron2}, 2019.

\bibitem[Xia et~al.(2024)Xia, Wang, Lv, Hao, and Shi]{xia2024vitcomer}
Chunlong Xia, Xinliang Wang, Feng Lv, Xin Hao, and Yifeng Shi.
\newblock {ViT-CoMer: Vision Transformer with Convolutional Multi-scale Feature Interaction for Dense Predictions}.
\newblock In \emph{CVPR}, 2024.

\bibitem[Xie et~al.(2021)Xie, Wang, Yu, Anandkumar, Alvarez, and Luo]{xie2021segformer}
Enze Xie, Wenhai Wang, Zhiding Yu, Anima Anandkumar, Jose~M Alvarez, and Ping Luo.
\newblock {SegFormer: Simple and Efficient Design for Semantic Segmentation with Transformers}.
\newblock In \emph{NeurIPS}, 2021.

\bibitem[Yang et~al.(2024)Yang, Kang, Huang, Xu, Feng, and Zhao]{yang2024depthanything}
Lihe Yang, Bingyi Kang, Zilong Huang, Xiaogang Xu, Jiashi Feng, and Hengshuang Zhao.
\newblock {Depth Anything: Unleashing the Power of Large-Scale Unlabeled Data}.
\newblock In \emph{CVPR}, 2024.

\bibitem[Yu et~al.(2022)Yu, Wang, Qiao, Collins, Zhu, Adam, Yuille, and Chen]{yu2022kmaxdeeplab}
Qihang Yu, Huiyu Wang, Siyuan Qiao, Maxwell Collins, Yukun Zhu, Hartwig Adam, Alan Yuille, and Liang-Chieh Chen.
\newblock {k-means Mask Transformer}.
\newblock In \emph{ECCV}, 2022.

\bibitem[Zhai et~al.(2022)Zhai, Kolesnikov, Houlsby, and Beyer]{zhai2022scaling}
Xiaohua Zhai, Alexander Kolesnikov, Neil Houlsby, and Lucas Beyer.
\newblock {Scaling Vision Transformers}.
\newblock In \emph{CVPR}, 2022.

\bibitem[Zhai et~al.(2023)Zhai, Mustafa, Kolesnikov, and Beyer]{zhai2023siglip}
Xiaohua Zhai, Basil Mustafa, Alexander Kolesnikov, and Lucas Beyer.
\newblock {Sigmoid Loss for Language Image Pre-Training}.
\newblock In \emph{ICCV}, 2023.

\bibitem[Zhao et~al.(2017)Zhao, Shi, Qi, Wang, and Jia]{zhao2017pspnet}
Hengshuang Zhao, Jianping Shi, Xiaojuan Qi, Xiaogang Wang, and Jiaya Jia.
\newblock {Pyramid Scene Parsing Network}.
\newblock In \emph{CVPR}, 2017.

\bibitem[Zhou et~al.(2017)Zhou, Zhao, Puig, Fidler, Barriuso, and Torralba]{zhou2017ade20k}
Bolei Zhou, Hang Zhao, Xavier Puig, Sanja Fidler, Adela Barriuso, and Antonio Torralba.
\newblock {Scene Parsing through ADE20K Dataset}.
\newblock In \emph{CVPR}, 2017.

\bibitem[Zhu et~al.(2021)Zhu, Su, Lu, Li, Wang, and Dai]{zhu2021deformabledetr}
Xizhou Zhu, Weijie Su, Lewei Lu, Bin Li, Xiaogang Wang, and Jifeng Dai.
\newblock {Deformable DETR: Deformable Transformers for End-to-End Object Detection}.
\newblock In \emph{ICLR}, 2021.

\end{thebibliography}
}

\clearpage
\appendix
\normalsize

\setcounter{table}{0}
\setcounter{figure}{0}

\renewcommand{\thetable}{\Alph{table}}
\renewcommand{\thesubtable}{\arabic{subtable}} %

\renewcommand{\thefigure}{\Alph{figure}}
\renewcommand{\thesubfigure}{\arabic{subfigure}} %

\appendix
\section*{Appendix}
\label{appendix}
\paragraph{Table of contents:}
\begin{itemize}[itemsep=-1pt,topsep=-1pt]
\item \S\ref{sec:supp:impl_details}: Implementation Details
\item \S\ref{sec:supp:exp}: Detailed Experimental Analysis
\item \S\ref{sec:supp:ood}: Out-of-distribution Confidence Estimation
\item \S\ref{sec:supp:qualitative}: Qualitative Examples
\end{itemize}

\section{Implementation Details}
\label{sec:supp:impl_details}
\subsection{Models}
\begin{figure}
    \centering
    \includegraphics[width=\linewidth]{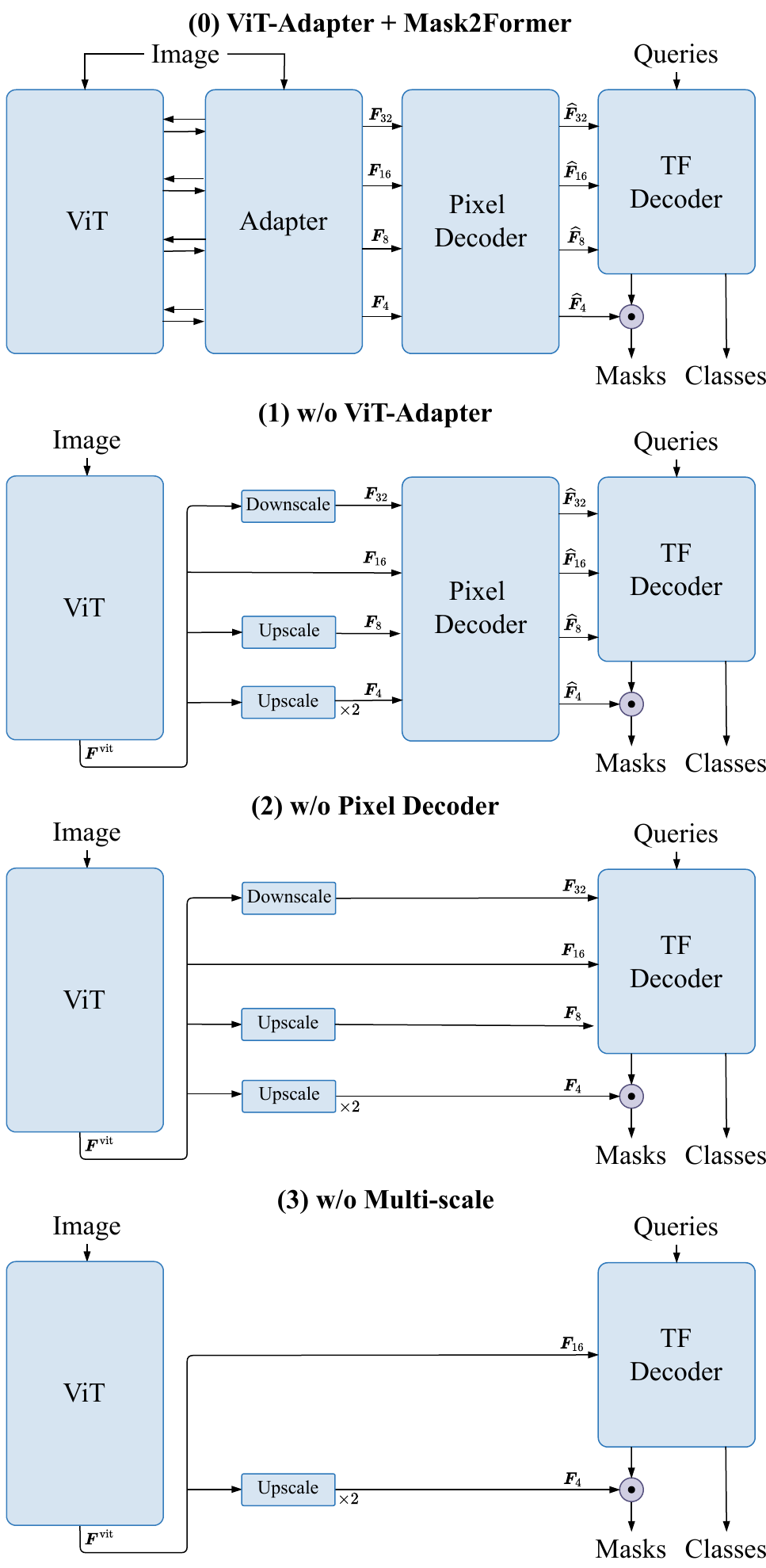}
    \caption{\textbf{Removing task-specific components.} We visualize the architectures of the resulting intermediate configurations.}
    \label{fig:supp:steps}
\end{figure}

\PARbegin{Visualizations of model configurations.}
In Sec.~3.2,
we explain how we gradually remove task-specific components. We visualize the architectures of the resulting intermediate configurations in~\cref{fig:supp:steps}. Here, the subscript $\bm{F}_i$ indicates that the features have a resolution of $\frac{1}{i}$ of the input image. The visualized model numbers correspond to those reported in 
\cref{tab:steps}.

\PAR{Libraries.}
For Mask2Former~\cite{cheng2022mask2former}, we use the implementation of \texttt{Huggingface Transformers}~\cite{wolf-etal-2020-transformers}. For pre-trained models, we use \texttt{timm}~\cite{rw2019timm}. 

\PAR{Pre-trained models.}
In \cref{tab:supp:timm_model}, we specify the \texttt{timm} model weights that we use for the experiments in this work. To support a patch size of $16\times16$ and different input sizes, we resize the patch embedding kernel and positional embeddings of pre-trained models following the FlexiViT~\cite{beyer2023flexivit} implementation of \texttt{timm}. Specifically, the patch embedding kernel is resized to a $16\times16$ patch size by approximately inverting the effect of patch resizing. The positional embeddings are resized to the required token grid size by using bicubic interpolation. The patch embedding kernel and positional embeddings are resized prior to fine-tuning, and keep the same size during fine-tuning.

\begin{table*}
\setlength{\tabcolsep}{30pt}
\centering
\footnotesize
\begin{tabularx}{1.0\linewidth}
{@{\hskip 2pt}Xll}
\toprule
Model & Pre-training & \texttt{timm} model \\
\midrule  

ViT-g & DINOv2~\cite{oquab2023dinov2,darcet2024registers} & \texttt{vit\_giant\_patch14\_reg4\_dinov2} \\

ViT-L & DINOv2~\cite{oquab2023dinov2,darcet2024registers} & \texttt{vit\_large\_patch14\_reg4\_dinov2} \\

ViT-B & DINOv2~\cite{oquab2023dinov2,darcet2024registers} & \texttt{vit\_base\_patch14\_reg4\_dinov2} \\

ViT-S & DINOv2~\cite{oquab2023dinov2,darcet2024registers} & \texttt{vit\_small\_patch14\_reg4\_dinov2} \\

ViT-L & EVA-02~\cite{fang2024eva02} & \texttt{eva02\_large\_patch14\_224.mim\_m38m} \\

ViT-L & DeiT-III (ImageNet-21K)~\cite{touvron2022deit,deng2009imagenet} & \texttt{deit3\_large\_patch16\_384.fb\_in22k\_ft\_in1k} \\

ViT-L & DeiT-III (ImageNet-1K)~\cite{touvron2022deit,deng2009imagenet} & \texttt{deit3\_large\_patch16\_384.fb\_in1k} \\

\bottomrule
\end{tabularx}
\caption[Model specification.]{\raggedright \textbf{Model specification.} For each ViT backbone~\cite{dosovitskiy2021vit} used in this work, we specify the \texttt{timm} model~\cite{rw2019timm} that we use.}
\label{tab:supp:timm_model}
\end{table*}

\begin{table*}
    \centering
    \footnotesize
    \setlength{\tabcolsep}{1.4pt}

    \begin{tabularx}{\linewidth}{
        c
        lcYY
        c
        YYY
        c
        YYYY
        }
        \toprule

        & \multirow{2}[2]{*}{Method} & \multirow{2}[2]{*}{Params} & \multirow{2}[2]{*}{GFLOPs} & \multirow{2}[2]{*}{FPS} &&
        \multicolumn{3}{c}{Panoptic Quality (PQ)} && \multicolumn{4}{c}{Average Precision (AP)} \\
        \cmidrule{7-9} \cmidrule{11-14}
        
        & & & & & &
        All & Things & Stuff && All & Large & Medium & Small \\
        
        \midrule

        \tablestep{0} & 
        ViT-Adapter + Mask2Former &
        349M &
        830 & 
        29 &&
        57.1 & 62.7 & 48.7 &&
        47.6 & 73.2 & 53.4 & 23.4 \\

        \midrule

        \tablestep{1} &
        \downrightarrow w/o ViT-Adapter & 
        342M & 
        700 & 
        36 &&
        56.7 & 62.3 & 48.3 &&
        46.9 & 72.7 & 52.9 & 22.7 \\

        \tablestep{2} &
        \hspace{4pt}\downrightarrow w/o Pixel decoder & 
        337M & 
        685 & 
        62 &&
        56.9 & 62.3 & 48.6 &&
        46.8 & 73.1 & 52.6 & 22.1 \\

        \tablestep{3} & 
        \hspace{8pt}\downrightarrow  w/o Multi-scale & 
        328M & 
        673 & 
        64 &&
        56.7 & 62.2 & 48.4 &&
        46.2 & 73.1 & 52.3 & 21.4 \\

        \tablestep{4} & 
        \hspace{12pt}\downrightarrow w/o Transformer decoder & 
        316M & 
        828 & 
        61 &&
        56.2 & 61.4 & 48.4 &&
        45.6 & 72.1 & 51.4 & 20.8 \\

        \tablestep{5} & 
        \hspace{16pt}\downrightarrow w/o Masking = \textbf{EoMT} &
        316M & 
        669 & 
        128
        &&
        56.0 & 61.2 & 48.2 &&
        45.2 & 72.2 & 51.0 & 20.3 \\
        \bottomrule

        \end{tabularx}
\caption{\textbf{From ViT-Adapter + Mask2Former to EoMT in detail.} Evaluated on COCO \textit{val2017}.}
\label{tab:supp:steps_detail}
\end{table*}

\PAR{Queries.}
In accordance with Mask2Former~\cite{cheng2022mask2former}, the models for panoptic and instance segmentation use $K = 200$ queries, while the models for semantic segmentation use $K = 100$ queries. For ViT-S and ViT-B we use $L_2 = 3$, for ViT-L we use $L_2 = 4$, and for ViT-g we use $L_2 = 5$. For EoMT, adding 200 tokens to a model that processes $640\times640$ images with a $16\times16$ patch size results in an increase of $12.5\%$ of the tokens processed by a ViT block, \textit{but only for the last $L_2$ ViT blocks}. As $L_1 = 20$ and $L_2 = 4$ for ViT-L, the total number of tokens processed in the entire ViT increases by only $2.1\%$.

\subsection{Training}
\label{sec:supp:training_details}

\PARbegin{Augmentation.}
During training, we apply the same data augmentation techniques as used by Mask2Former~\cite{cheng2022mask2former}. Specifically, training images undergo random horizontal flipping, random scale jittering, padding if necessary, and random cropping. Random color jittering is additionally applied for ADE20K~\cite{zhou2017ade20k} and Cityscapes~\cite{cordts2016cityscapes}. For panoptic and instance segmentation, we use large-scale jitter~\cite{ghiasi2021copypaste} (between $0.1\times$ and $2.0\times$), and for semantic segmentation we use normal-scale jitter (between $0.5\times$ and $2.0\times$). 

\PAR{Loss function.}
To supervise our models, we adopt the same loss function as Mask2Former~\cite{cheng2022mask2former}. Specifically, across all tasks and datasets, we use the cross-entropy (CE) loss for the class logits, and the binary-cross entropy (BCE) and the Dice loss~\cite{milletari2016vnet} for the mask logits. The individual losses are weighted using scalars, resulting in the total loss function:
\begin{equation} \mathcal{L}_{\textrm{tot}} = \lambda_{\textrm{bce}} \mathcal{L}_{\textrm{bce}} + \lambda_{\textrm{dice}} \mathcal{L}_{\textrm{dice}} + \lambda_{\textrm{ce}} \mathcal{L}_{\textrm{ce}}, \end{equation}
where $\lambda_{\textrm{bce}}$, $\lambda_{\textrm{dice}}$, and $\lambda_{\textrm{ce}}$ are set to 5.0, 5.0, and 2.0, respectively, following Mask2Former~\cite{cheng2022mask2former}.

\PAR{Learning rate warm-up.}
We use a two-stage linear learning rate warm-up for all models. In practice, we first warm-up the randomly initialized parameters for 500 iterations, while keeping the pre-trained parameters frozen. After 500 iterations, we warm-up the pre-trained parameters for 1000 iterations. In both cases, the initial learning rate is set to 0.

\subsection{Evaluation}
\label{sec:supp:eval_details}
\PARbegin{Image processing.}
For panoptic and instance segmentation, we use padded inference, resizing the longer side of the image to the input size, and padding the shorter side with zeros to create a square image. For semantic segmentation, we apply windowed inference, resizing the shorter side of the image to the input size, and processing the image through the model in several proportionally spaced square crops, in a sliding-window manner~\cite{kerssies2024benchmark}.

\PAR{Efficiency measurements.}
For existing works, we report FLOPs from the respective papers but measure FPS the same way that we measure it for our models, on the same hardware. For ViT-Adapter + M2F and our models, we calculate the FLOPs ourselves.
When measuring FPS, \texttt{torch.compile}~\cite{ansel2024pytorch2} is disabled for Mask2Former~\cite{cheng2022mask2former} with Swin-L~\cite{liu2021swin} on ADE20K~\cite{zhou2017ade20k} due to compilation errors. On COCO~\cite{lin2014coco}, \texttt{torch.compile} only yields a small speedup for this model ($<10\%$). Additionally, mixed precision is not supported for OneFormer~\cite{jain2023oneformer} with DiNAT-L~\cite{hassani2022dinat}, thus we use full precision here.

\PAR{Token merging.}
For our token merging experiment in 
\cref{sec:exp:further} and~\cref{tab:algm},
we evaluate the throughput of the model in \textit{images per second}, following existing work for token merging~\cite{norouzi2024algm,bolya2023tome}. This means that we use a batch size of 32, apply ALGM~\cite{norouzi2024algm} for token merging, and report the number of images that are processed per second, averaged over the entire validation set. ALGM adaptively determines the number of tokens that should be merged per image, based on image complexity. To allow batch processing, we identify the lowest number of mergeable tokens per image across the batch according to the ALGM token merging criterion, and use that number of merged tokens for all images in the batch. 

Importantly, ALGM is applied only during inference. Thus, the throughput improvement in \cref{tab:algm}
is achieved simply by applying ALGM to EoMT and processing batches of images, with no additional training required.

\begin{figure*}
    \centering
    \begin{subfigure}[b]{.32\linewidth}
         \centering
         \includegraphics[width=\linewidth]{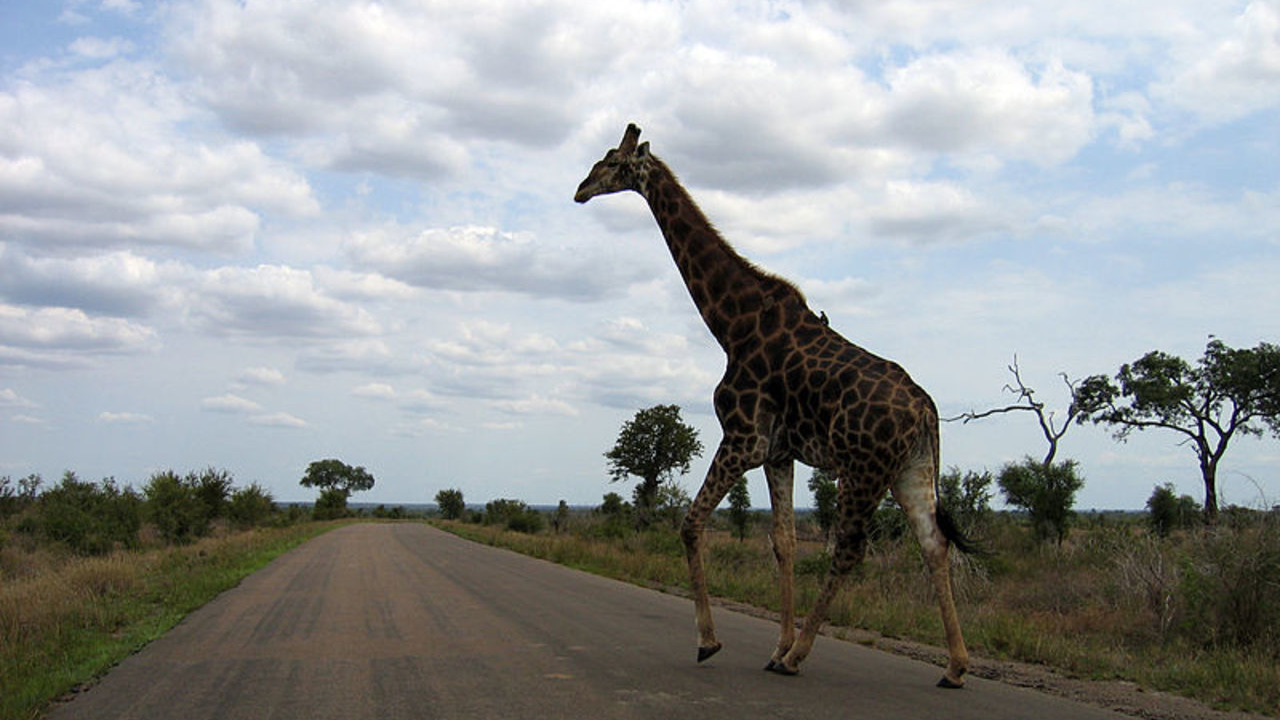}
         \caption{Input image}
         \label{fig:sub:input}
    \end{subfigure}
    \begin{subfigure}[b]{.32\linewidth}
         \centering
         \includegraphics[width=\linewidth]{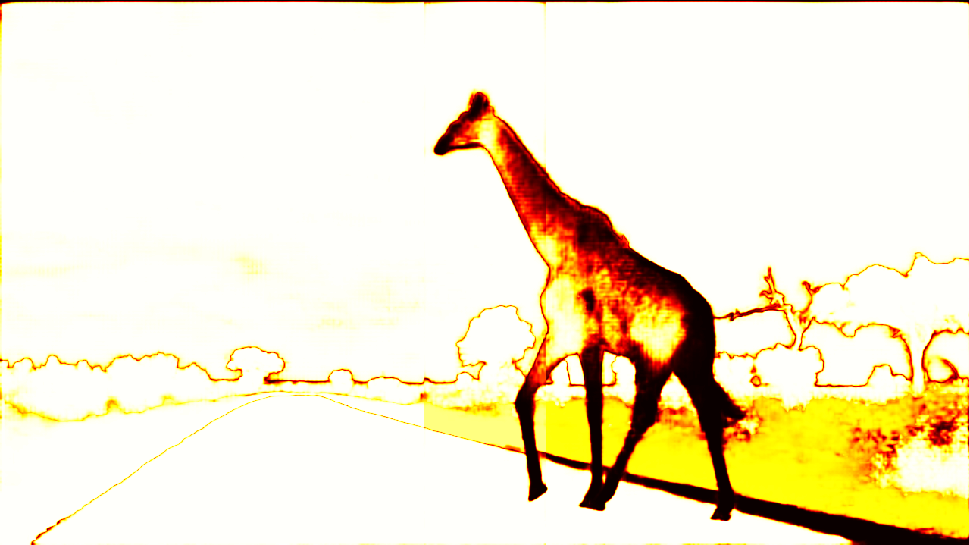}
         \caption{ViT-Adapter + M2F~\cite{chen2023vitadapter,cheng2022mask2former} confidence}
         \label{fig:sub:vitadapter}
    \end{subfigure}
    \begin{subfigure}[b]{.32\linewidth}
         \centering
         \includegraphics[width=\linewidth]{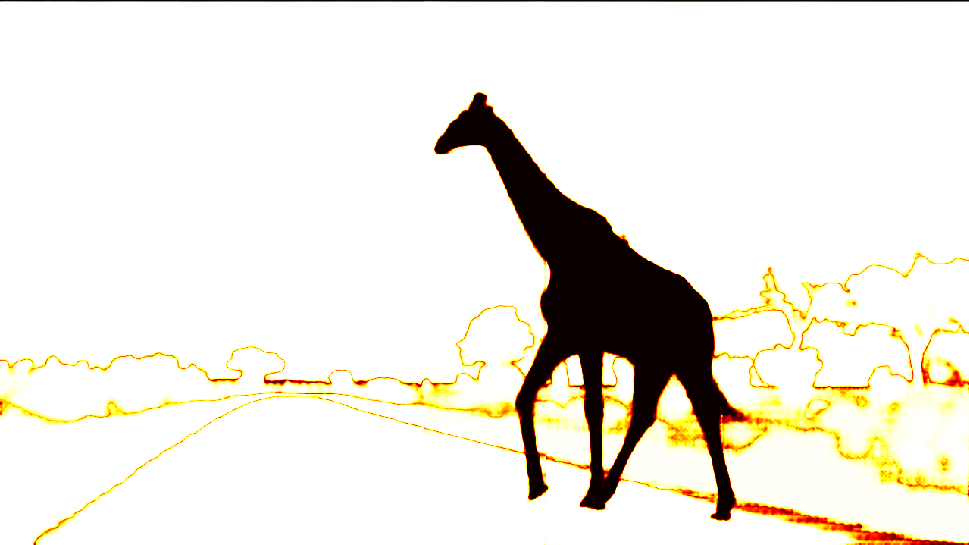}
         \caption{\textbf{EoMT} (Ours) confidence}
         \label{fig:sub:eomt}
    \end{subfigure}
    \caption{\textbf{Qualitative comparison of out-of-distribution (OOD) confidence estimation.} EoMT reliably assigns low confidence to the full OOD object, while ViT-Adapter + M2F only does so partially. Darker colors indicate lower confidence. Trained on Cityscapes~\textit{train}~\cite{cordts2016cityscapes}, evaluated on BRAVO~\cite{vu2024bravo}.}
    \label{fig:supp:ood}
\end{figure*}

\section{Detailed Experimental Analysis}
\label{sec:supp:exp}

\PARbegin{From ViT-Adapter + M2F to EoMT in detail.}
In \cref{tab:supp:steps_detail}, we provide more detailed results on the impact of the removal of task-specific components on both panoptic and instance segmentation on COCO~\cite{lin2014coco}. For panoptic segmentation, we not only report the overall Panoptic Quality (PQ), but also separately the PQ for countable thing classes (PQ\textsuperscript{th}) and uncountable stuff classes (PQ\textsuperscript{st}). Similarly, for instance segmentation, we separately report AP for large (AP\textsuperscript{L}), medium (AP\textsuperscript{M}), and small objects (AP\textsuperscript{S}).

\PAR{General applicability of mask annealing.}
In \cref{tab:supp:mask_annealing_detail}, we assess the effect of our mask annealing strategy for both EoMT and the ViT-Adapter + M2F baseline. The results demonstrate the general applicability of mask annealing, as it is also effective for ViT-Adapter + M2F.

\begin{table}[t]
\setlength{\tabcolsep}{1.5pt}
\footnotesize
\centering
\begin{tabularx}{\linewidth}{cl @{\hskip 5pt}cl c Yc}
\toprule
\multicolumn{2}{l}{\multirow{2}[1]{*}{Training}} & \multicolumn{2}{l}{\multirow{2}[1]{*}{Inference}} &~~~& \multicolumn{2}{c}{Panoptic Quality (PQ)} \\
\cmidrule{6-7}
\multicolumn{2}{l}{} & \multicolumn{2}{l}{} & & EoMT & ViT-Ad.~+ M2F \\
\midrule

\rcmark & Masking &  %
\rcmark & Masking &&  %
\phantom{\qdecnew{0.00}}56.2\phantom{\qdecnew{0.00}} &
\phantom{\qdecnew{0.00}}57.1\phantom{\qdecnew{0.00}}  \\

\gxmark & w/o Masking &  %
\gxmark & w/o Masking &&  %
\phantom{\qdecnew{00.0}}53.2\qdecnew{3.0\phantom{0}} &
\phantom{\qdecnew{18.1}}54.0\qdecnew{3.1\phantom{0}}  \\

\rcmark \fancyrightarrow \gxmark&\textit{Mask annealing} &  %
\gxmark & w/o Masking &&  %
\phantom{\qdecnew{00.0}}56.0\qdecnew{0.2\phantom{0}} &
\phantom{\qdecnew{00.0}}56.8\qdecnew{0.3\phantom{0}}  \\

\bottomrule
\end{tabularx}
\caption{\textbf{\textit{Mask annealing}}. Effective for both EoMT and ViT-Adapter + M2F~\cite{chen2023vitadapter,cheng2022mask2former}. When never masking, intermediate masks are not predicted or supervised. Evaluated on COCO \textit{val2017}~\cite{lin2014coco}.}%
\label{tab:supp:mask_annealing_detail}
\end{table}

\begin{table}[t]
\setlength{\tabcolsep}{10pt}
\centering
\footnotesize
\begin{tabularx}{1.0\linewidth}
{cYYYY}
\toprule
\# Blocks ($L_2$) & Params & GFLOPs & FPS & PQ \\
\midrule  
9      & 316 & 688 & 126 & 55.7 \\
6      & 316 & 676 & 127 & 55.7 \\
4      & 316 & 669 & 128 & \textbf{56.0} \\
2      & 316 & 660 & 128 & 55.4 \\ 
\bottomrule
\end{tabularx}
\caption{\textbf{Number of blocks that process queries.} The model with $L_2=4$ achieves the best PQ, while FPS is not significantly affected by changing $L_2$. Evaluated on COCO \textit{val2017}~\cite{lin2014coco}.}%
\label{tab:supp:blocks}
\end{table}

\PAR{Number of blocks that process queries.}
In~\cref{tab:supp:blocks}, we examine the impact of varying $L_2$, \ie, the number of ViT blocks in EoMT that process queries as well as patch tokens. EoMT demonstrates stable performance across different configurations, with the highest PQ for ViT-L observed around $L_2=4$, while the prediction speed in FPS is not significantly affected by changing $L_2$. Consequently, we set $L_2=4$ as the default configuration for ViT-L.

\section{Out-of-distribution Confidence Estimation}
\label{sec:supp:ood}
In \cref{sec:exp:further},
we discuss the out-of-distribution (OOD) generalization capabilities of EoMT. There, we show that DINOv2-based models, such as EoMT, significantly outperform non-ViT-based models such as Swin~\cite{liu2021swin} in OOD generalization despite similar in-distribution (ID) performance. 

\begin{table}
    \centering
    \footnotesize
    \setlength{\tabcolsep}{2.0pt}
    \begin{tabularx}{\linewidth}{
    llcY
    }
    \toprule

    Method & Backbone & Pre-training & $\text{AUPRC}\textsubscript{OOD}$ \\

    \midrule

    M2F~\cite{cheng2022mask2former} &
    Swin-L~\cite{liu2021swin} &
    IN21K &
    56.8 \\

    M2F$^\ddagger$~\cite{cheng2022mask2former} &
    ViT-Adapter-L$^\ddagger$~\cite{chen2023vitadapter} &
    DINOv2 &
    68.7 \\

    \textbf{EoMT} (Ours) &
    ViT-L~\cite{dosovitskiy2021vit} &
    DINOv2 &
    \textbf{89.7} \\

    \bottomrule
    
    \end{tabularx}
    \caption{\textbf{Quantitative comparison of out-of-distribution (OOD) confidence estimation.} EoMT achieves the highest AUPRC\textsubscript{OOD}, demonstrating its superior confidence estimation. Trained on Cityscapes~\textit{train}~\cite{cordts2016cityscapes}, evaluated on BRAVO~\cite{vu2024bravo}. $^\ddagger$Our re-implementation.}
    \label{tab:supp:ood}
\end{table}

Next, we also assess how well different models distinguish OOD regions from ID regions with their confidence scores. OOD regions, as defined in the BRAVO~\cite{vu2024bravo} benchmark, refer to novel object classes that were not present in the training data. We report the AUPRC\textsubscript{OOD} metric, which quantifies the model’s ability to assign lower confidence to these unseen objects, ensuring they can be correctly identified as OOD.

As shown in~\cref{tab:supp:ood}, EoMT achieves an AUPRC\textsubscript{OOD} of 89.7, significantly outperforming ViT-Adapter + M2F~\cite{chen2023vitadapter,cheng2022mask2former} with a score of 68.7 and Swin~\cite{liu2021swin} + M2F with a score of 56.8. The visualization in~\cref{fig:supp:ood} further highlights that EoMT consistently assigns low confidence to the OOD object while maintaining high confidence for ID regions. In contrast, ViT-Adapter + M2F~\cite{chen2023vitadapter,cheng2022mask2former} fails to reliably assign low confidence to all OOD pixels.

\section{Qualitative Examples}
\label{sec:supp:qualitative}
\begin{figure*}[p]
    \centering

    \begin{subfigure}[b]{\linewidth}
        \centering
        \includegraphics[height=3.25cm]{images/input_image_0.png}
        \includegraphics[height=3.25cm]{images/input_image_437.png}
        \includegraphics[height=3.25cm]{images/input_image_615.png}
        \includegraphics[height=3.25cm]{images/input_image_3665.png}
        \caption{Input images}
    \end{subfigure}\\

    \begin{subfigure}[b]{\linewidth}
        \centering
        \includegraphics[height=3.25cm]{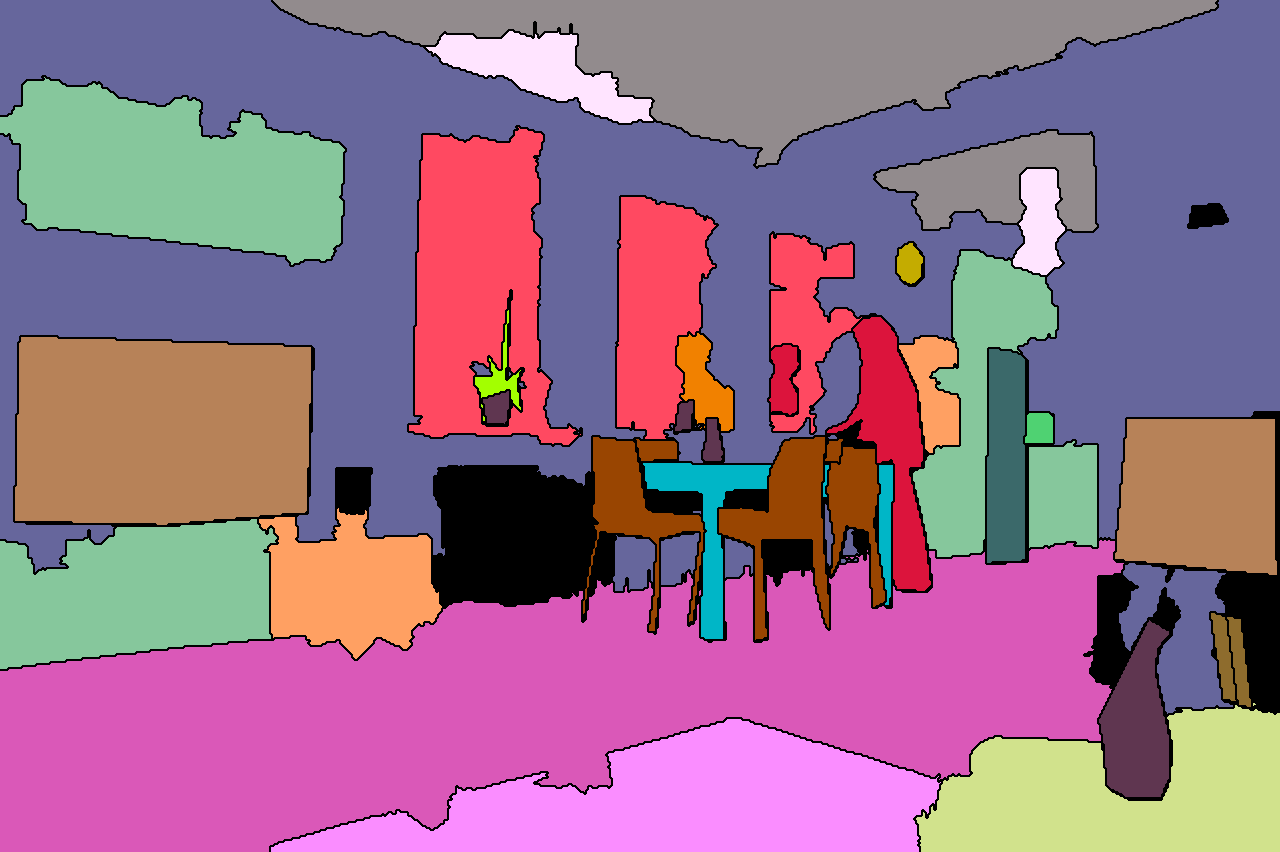}
        \includegraphics[height=3.25cm]{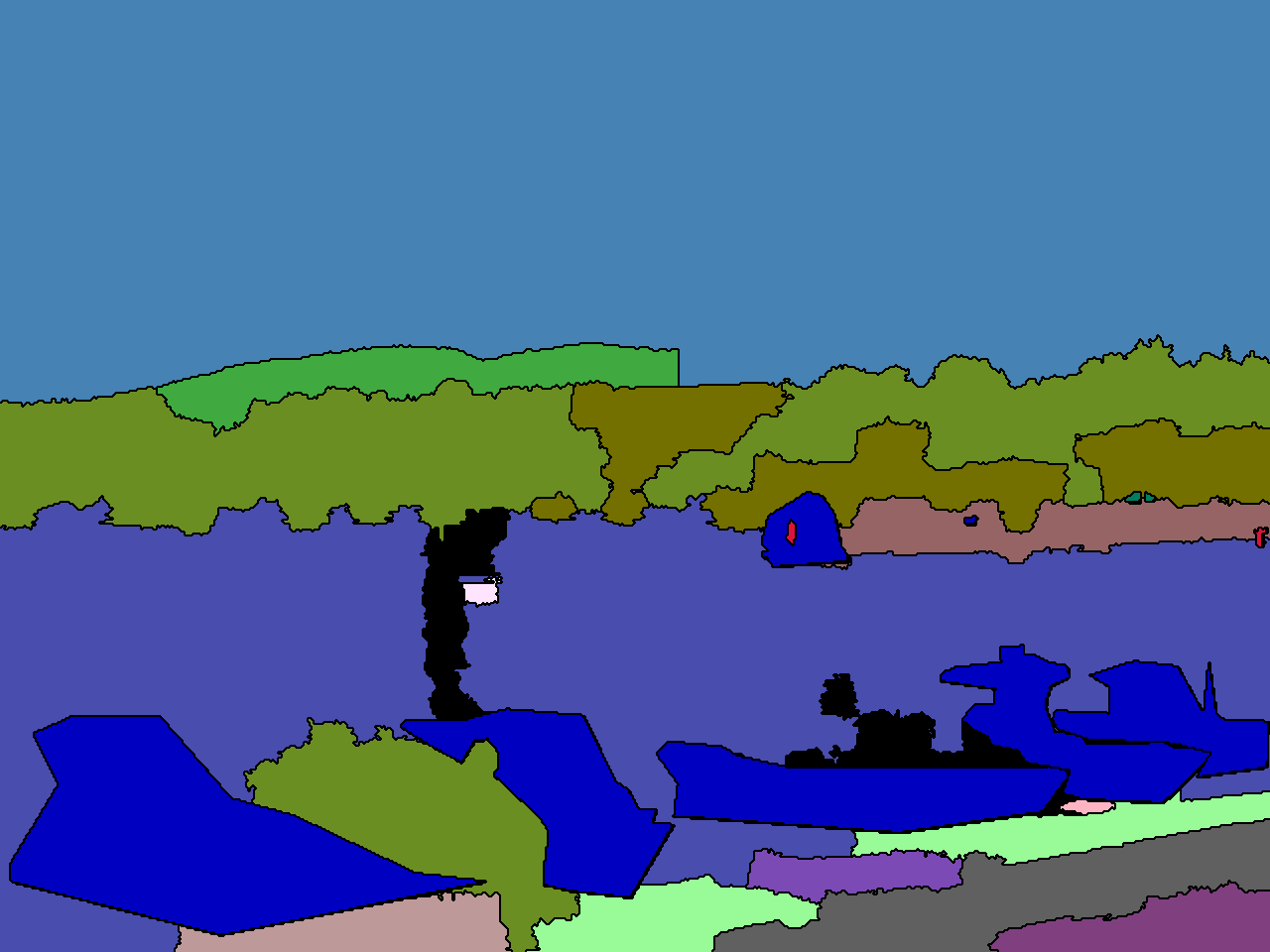}
        \includegraphics[height=3.25cm]{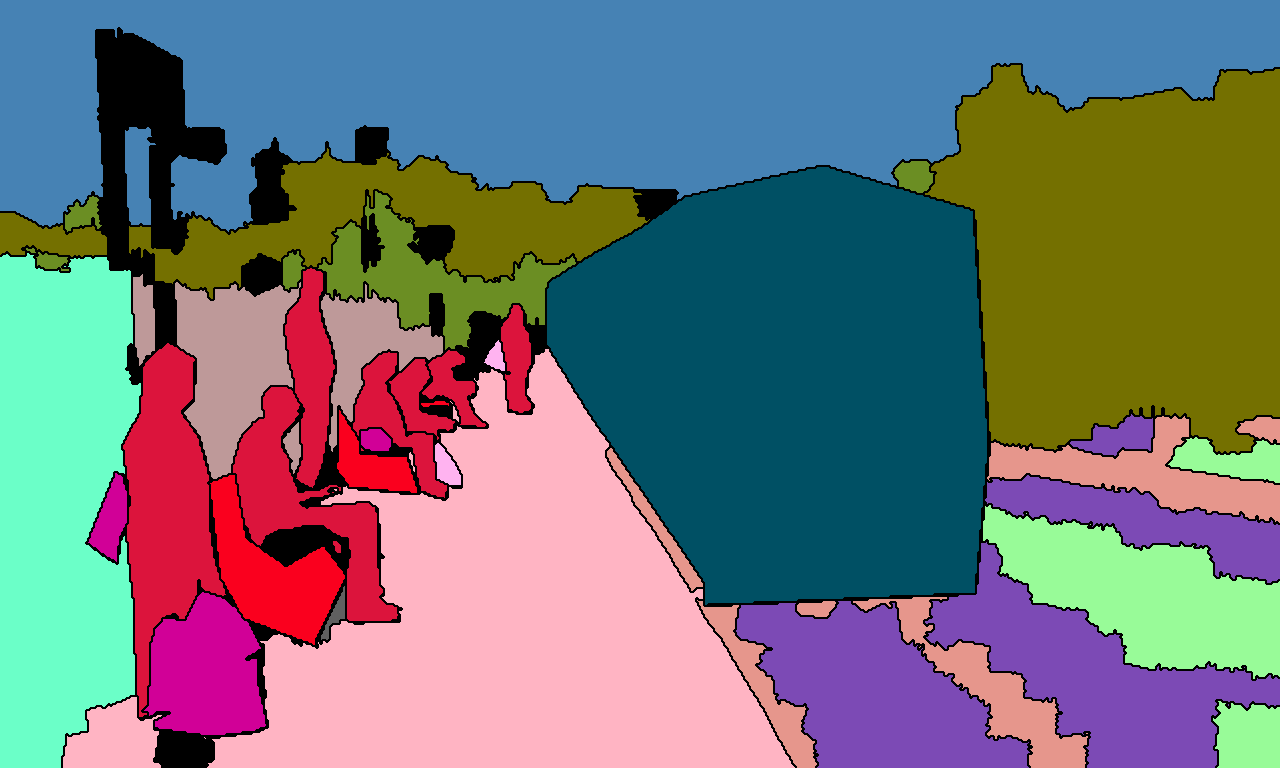}
        \includegraphics[height=3.25cm]{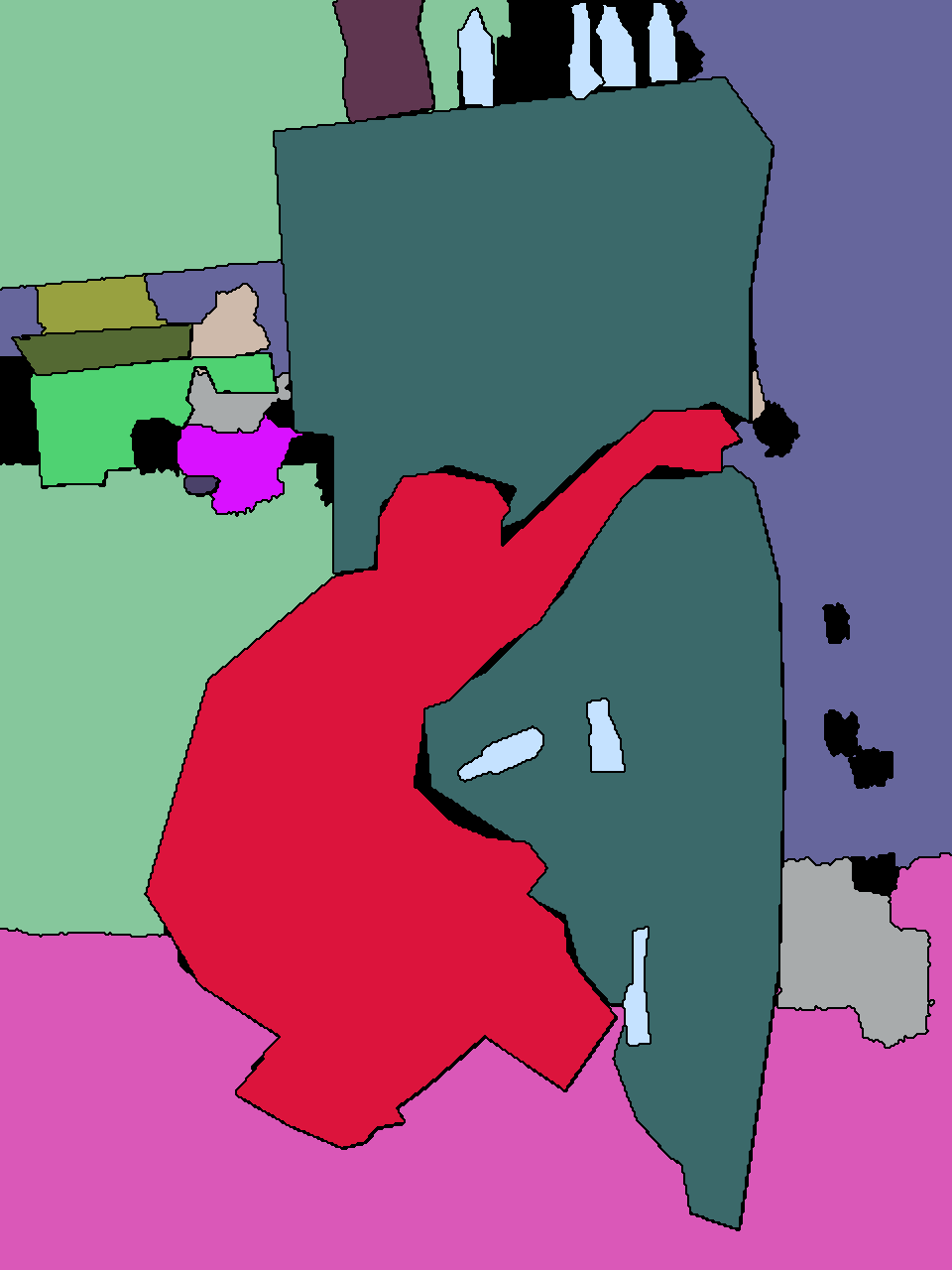}
        \caption{Ground-truth annotations}
    \end{subfigure}\\

    \begin{subfigure}[b]{\linewidth}
        \centering
        \includegraphics[height=3.25cm]{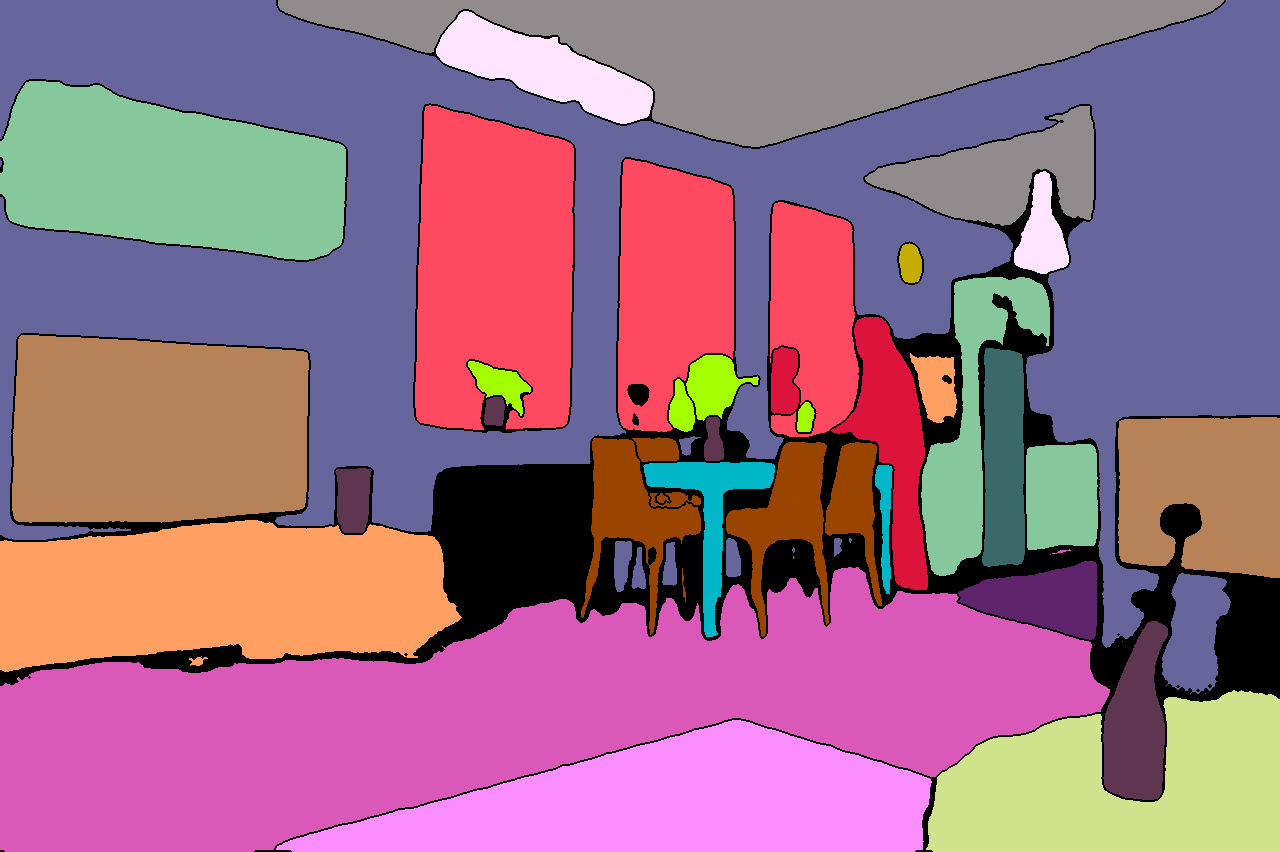}
        \includegraphics[height=3.25cm]{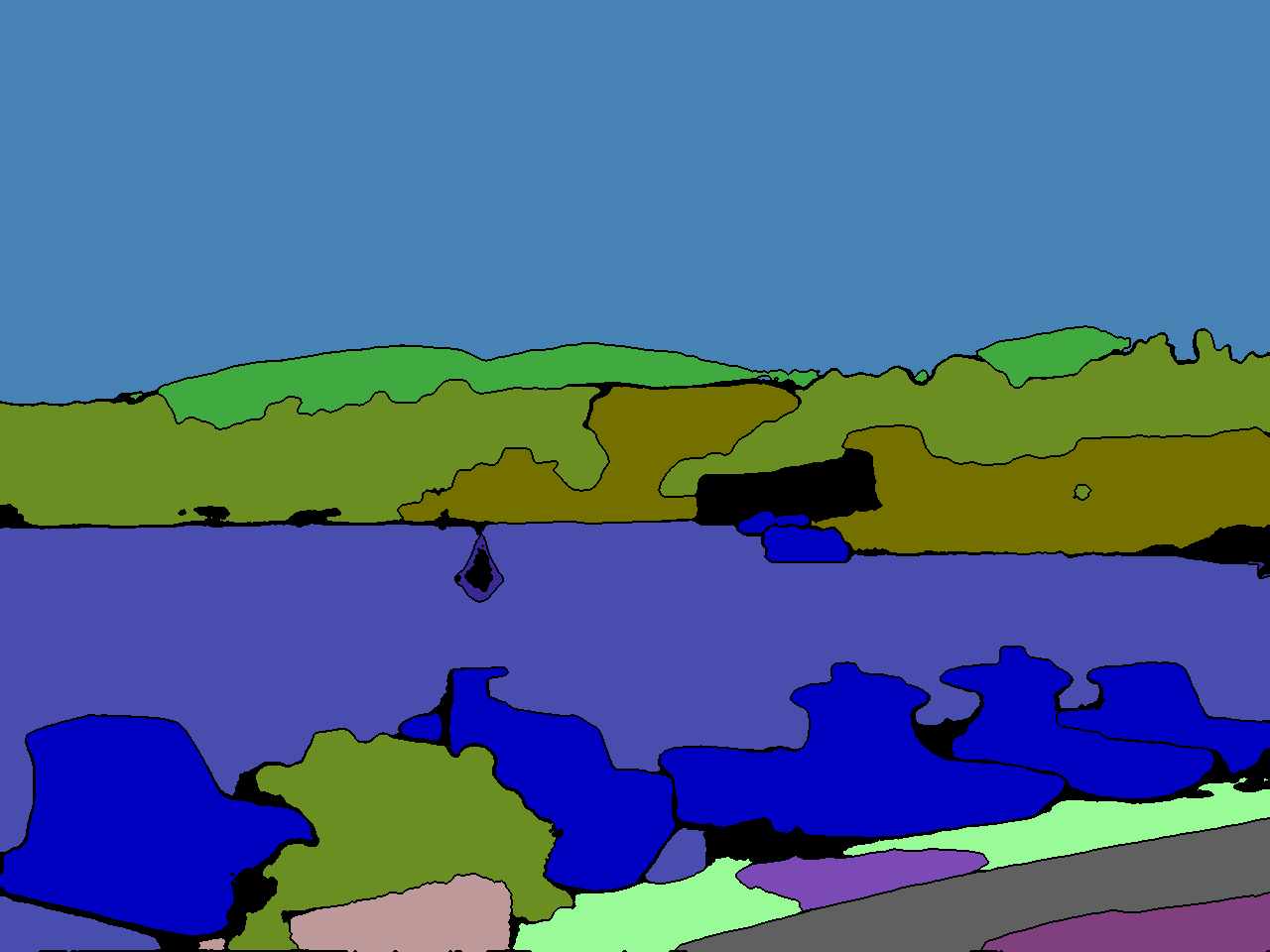}
        \includegraphics[height=3.25cm]{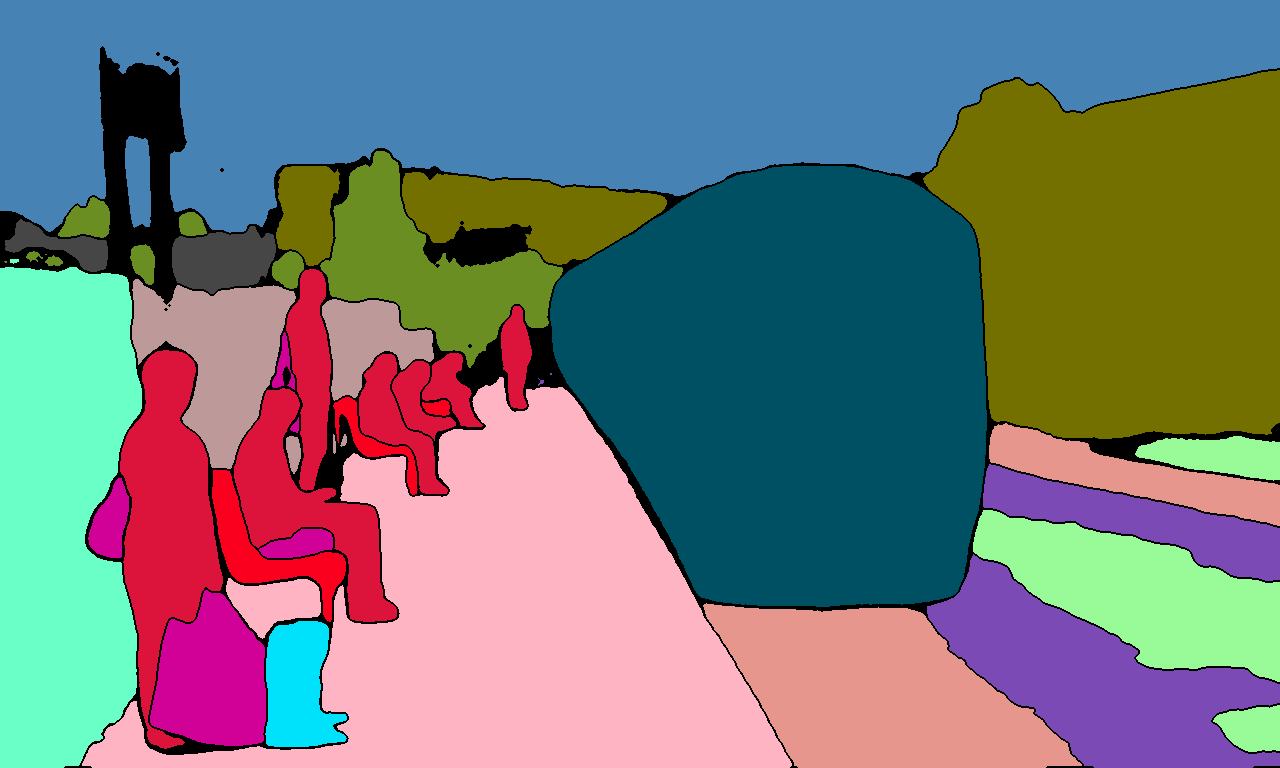}
        \includegraphics[height=3.25cm]{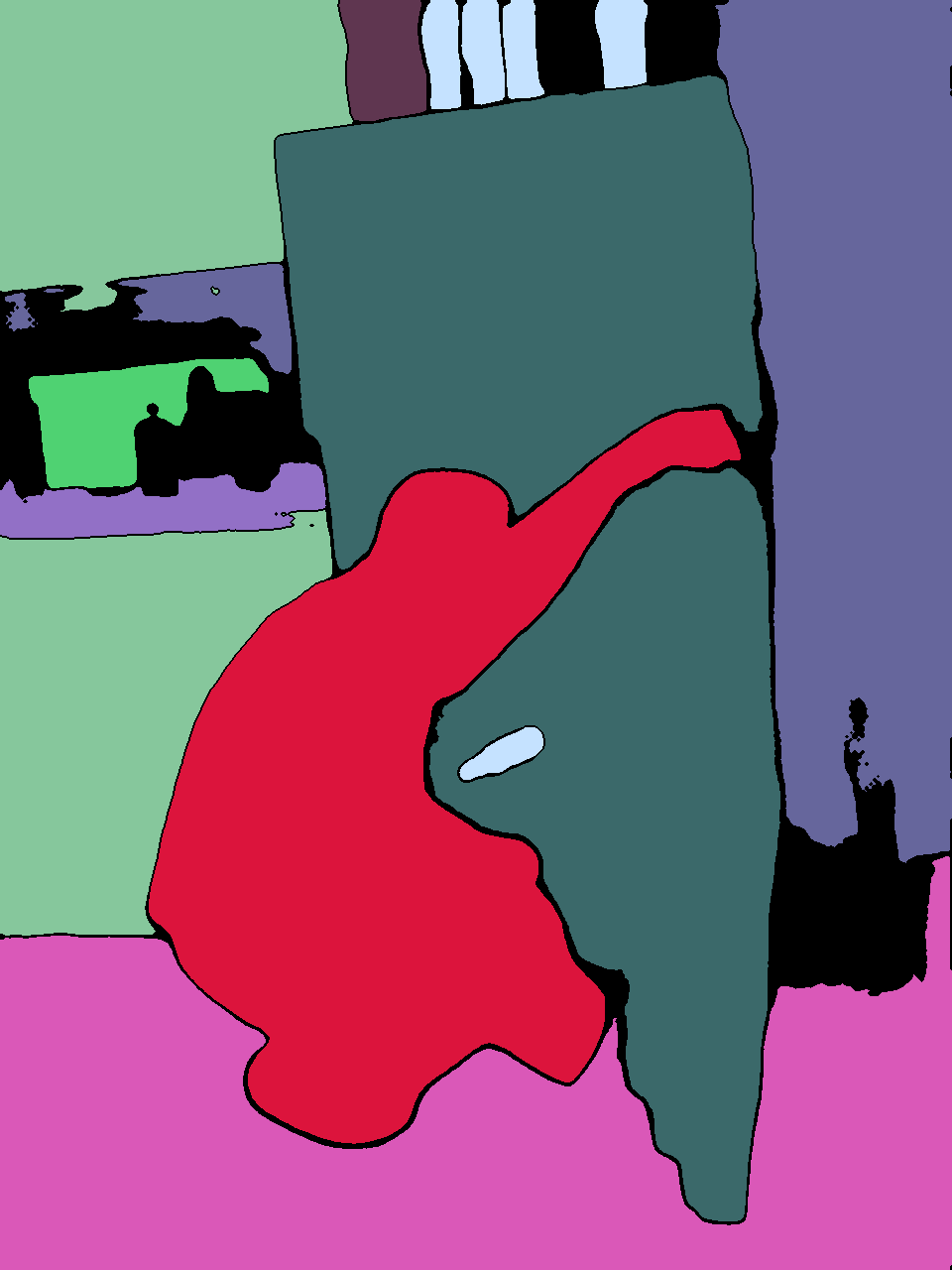}
        \caption{ViT-Adapter + M2F~\cite{chen2023vitadapter,cheng2022mask2former} predictions}
    \end{subfigure}\\

    \begin{subfigure}[b]{\linewidth}
        \centering
        \includegraphics[height=3.25cm]{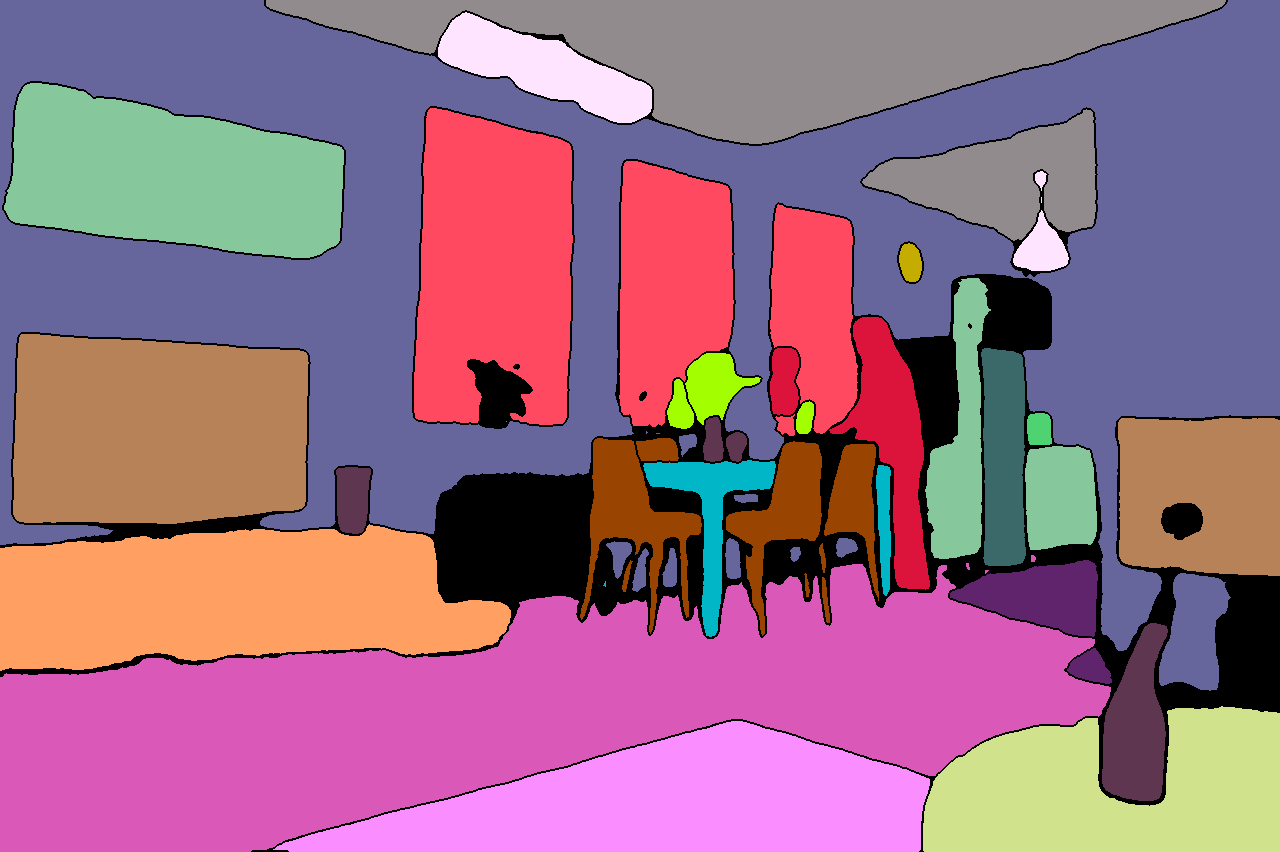}
        \includegraphics[height=3.25cm]{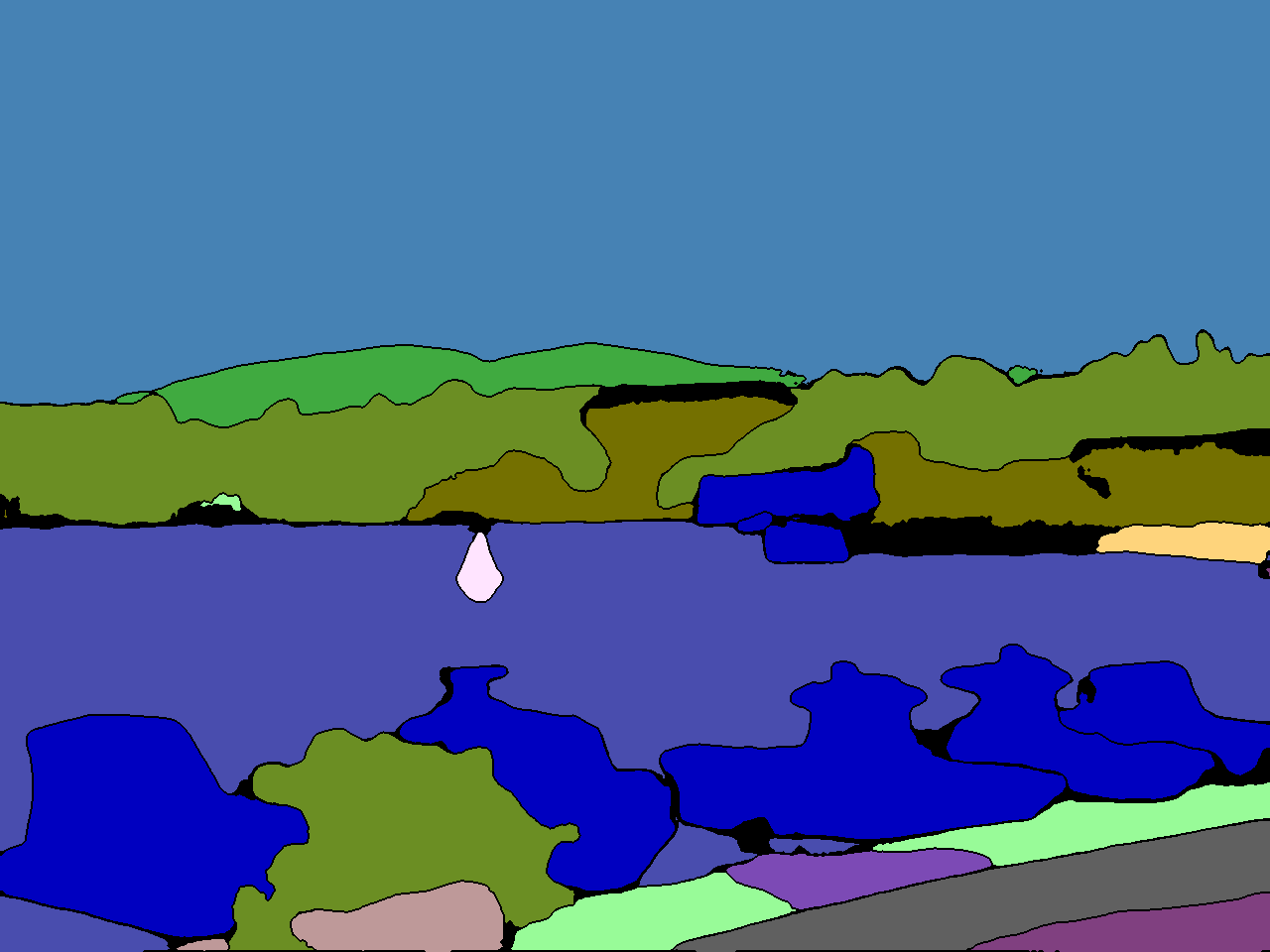}
        \includegraphics[height=3.25cm]{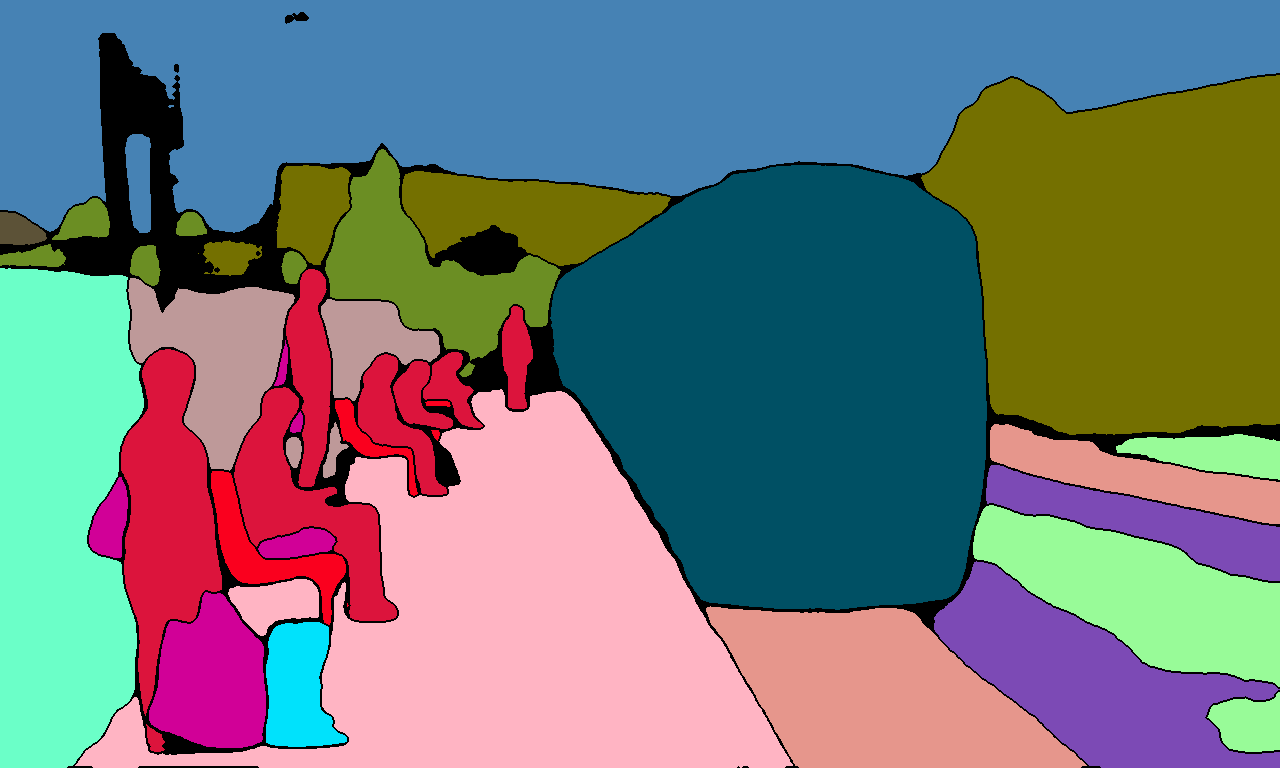}
        \includegraphics[height=3.25cm]{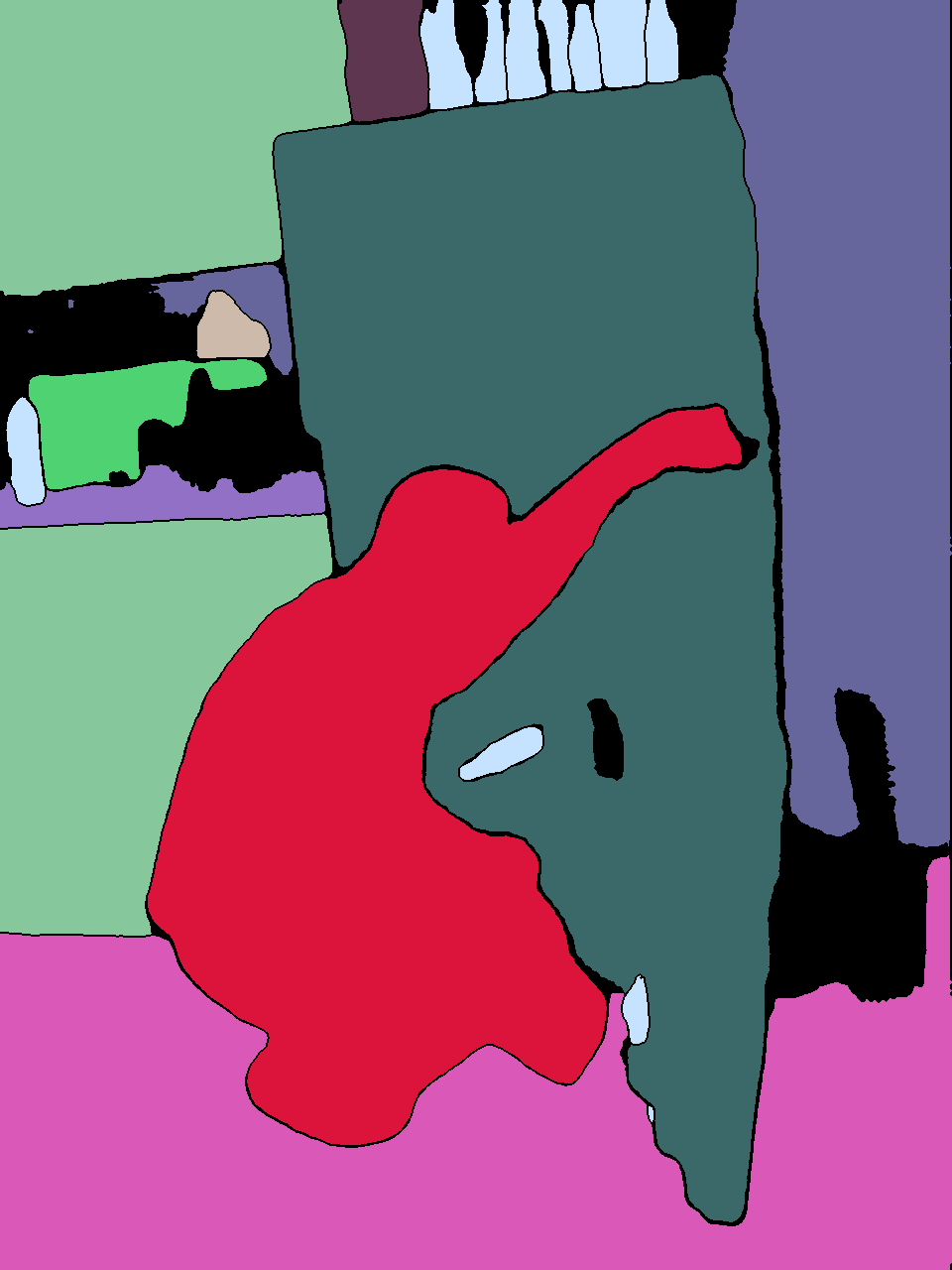}
        \caption{\textbf{EoMT} (Ours) predictions}
    \end{subfigure}

    \caption{\textbf{Qualitative examples for panoptic segmentation on COCO~\cite{lin2014coco}.} Using DINOv2-g~\cite{oquab2023dinov2} and a $1280\times1280$ input size.}
    \label{fig:supp:qualitative}
\end{figure*}

In~\cref{fig:supp:qualitative} we visualize predictions of ViT-Adapter + M2F~\cite{chen2023vitadapter,cheng2022mask2former} and EoMT for panoptic segmentation on COCO~\cite{lin2014coco}.

\end{document}